\documentclass[lettersize,journal]{IEEEtran}
\usepackage{amsmath,amsfonts}
\usepackage{algorithmic}
\usepackage{algorithm}
\usepackage{array}
\usepackage[caption=false,font=footnotesize]{subfig}
\usepackage{textcomp}
\usepackage{stfloats}
\usepackage{url}
\usepackage{verbatim}
\usepackage{graphicx}
\usepackage{cite}
\usepackage{makecell}
\usepackage{booktabs}
\usepackage{multirow}
\usepackage[hidelinks]{hyperref}
\usepackage{xcolor}
\usepackage{microtype}

\usepackage{bm}
\usepackage{enumitem}

\begin{document}

\title{Dynamic Model Identification and Gravity Compensation for the dVRK-Si Patient Side Manipulator}

\author{Haoying Zhou$^{1,2}$, Hao Yang$^{3}$, Brendan Burkhart$^{2}$, Anton Deguet$^{2}$, Zhonghao Zhang$^{3}$,\\ Loris Fichera$^{1}$, Gregory S. Fischer$^{1}$, Jie Ying Wu$^{3}$ and Peter Kazanzides$^{2,4}$
\thanks{This work was supported in part by NSF AccelNet awards OISE-1927275 and OISE-1927354.}
\thanks{$^{1}$Department of Robotics Engineering, Worcester Polytechnic Institute, MA 01609, USA. (Email: \textit{hzhou6@wpi.edu})}
\thanks{$^{2}$Laboratory for Computational Sensing and Robotics, Johns Hopkins University, MD 21218, USA.}
\thanks{$^{3}$Department of Computer Science, Vanderbilt University, TN 37212, USA.}
\thanks{$^{4}$Department of Computer Science, Johns Hopkins University, MD 21218, USA. (Email: \textit{pkaz@jhu.edu})}
}

\maketitle

\begin{abstract}

The da Vinci Research Kit (dVRK) has been widely used for research in robot-assisted surgery over the past decade, and its latest generation, the dVRK-Si was made publicly available in 2025. This manuscript presents the first complete kinematic and dynamic modeling framework for the dVRK-Si Patient-Side Manipulator (PSM), whose redesigned mechanism differs substantially from that of the first-generation dVRK and therefore requires a new formulation to accurately capture gravitational effects and improve overall control performance. We first develop a modified DH kinematic model that represents the dVRK-Si PSM's redesigned mechanism. The equations of motion are then derived using the Euler-Lagrange formulation, and the inverse dynamics are expressed in a linear-in-parameters regressor form. The dynamic parameters are identified from data collected along a periodic excitation trajectory optimized for numerical conditioning and estimated by convex optimization subject to the physical feasibility constraints. The identified model is integrated into the dVRK control stack to enable real-time gravity compensation and computed-torque feedforward control. Experiments on a physical dVRK-Si demonstrate that the gravity compensation reduces steady-state joint errors by up to 84\% and decreases end-effector tip drift during static holds from 4.2 mm to 0.7 mm. Computed-torque feedforward control further improves transient and trajectory-tracking accuracy. During sinusoidal trajectory tracking, computed-torque feedforward reduces position errors by 35\% versus gravity-only feedforward and by 40\% versus PID-only. The proposed pipeline could support reliable control, high-fidelity simulation, and learning-based automation on the dVRK-Si.

\end{abstract}


\section{Introduction}

\begin{figure}[!t]
    \centering
    \includegraphics[width=0.45\linewidth]{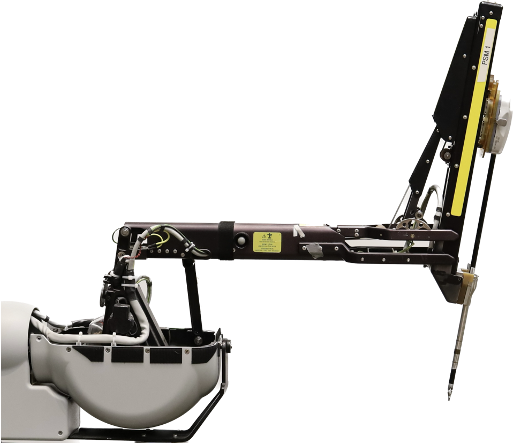}
    \hfill
    \includegraphics[width=0.45\linewidth]{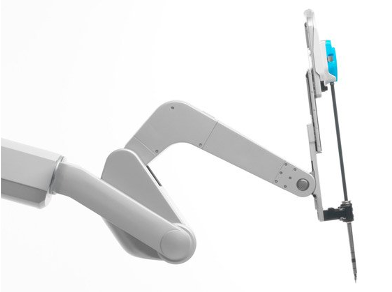}
    \caption{Patient Side Manipulator (PSM): dVRK Classic (left), dVRK-Si (right).}
    \label{fig:psm_diff}
\end{figure}

The da Vinci Research Kit (dVRK, also known as dVRK Classic)~\cite{kazanzides2014open} has become a widely adopted open-source platform for research in robot-assisted minimally invasive surgery, enabling reproducible studies in teleoperation, perception, learning, and autonomy~\cite{d2021accelerating, kim2025srt}. Recently, a new dVRK generation built from da Vinci Si hardware (dVRK-Si)~\cite{xu2025dvrk} has become available to the community. Compared with the first-generation da Vinci components used in the dVRK Classic, the dVRK-Si introduces a redesigned Patient Side Manipulator (PSM) structure that substantially changes the arm’s mass distribution and gravitational loading, as shown in Fig~\ref{fig:psm_diff}. As a result, gravity effects on the dVRK-Si PSM can no longer be treated as negligible, and unmodeled gravity produces noticeable steady-state errors and drift during static holds, degrading accuracy and responsiveness~\cite{yang2025effectiveness, zhou2025gravity} in both teleoperation and autonomous behaviors.

Model-based control and realistic simulation both depend on accurate robot dynamics. While dynamic modeling and identification methods have been explored for the dVRK Classic~\cite{fontanelli2017modelling, wang2019convex, argin2024davinci, yang2024hybrid}, differences in mechanical design and transmission mechanisms limit the transferability of previously identified parameters to the dVRK-Si. Moreover, the dVRK-Si platform creates a practical need for gravity compensation that is accurate enough to improve control performance, yet efficient and robust enough to integrate into the real-time control stack.

This paper presents a complete modeling and identification framework for the dVRK-Si PSM and demonstrates its impact on gravity compensation and dynamic position tracking. We first derive a full modified Denavit–Hartenberg (DH) kinematic model for the dVRK-Si PSM, including the closed-chain structure associated with the remote-center-of-motion (RCM) mechanism. We then formulate the robot dynamics using the Euler–Lagrange method and express the inverse dynamics in a linear-in-parameters form. Using an optimized excitation trajectory and a convex optimization-based identification procedure with physical feasibility constraints, we estimate a consistent set of dynamic parameters for a physical dVRK-Si system. With the identified parameters, we compute the gravity vector in real time and implement gravity compensation within the control loop. Furthermore, we compute the full dynamics in real time and add it as a feedforward term within the control loop.

We evaluate the proposed framework on a physical dVRK-Si PSM. Experimental results show that gravity compensation based on the identified model substantially improves static performance, reducing steady-state joint errors by \textcolor{red}{up to} 84\% and decreasing end-effector tip drift during static holds from 4.2\,mm to 0.7\,mm (computed via forward kinematics of the measured joint drift). Beyond these headline metrics, we provide a detailed analysis of torque bounds, trajectory-level identification accuracy, and practical considerations for integrating gravity compensation into the dVRK software stack. For dynamic trajectory tracking, adding the computed-torque feedforward reduces position tracking errors for the first three joints by about 35\% compared to gravity feedforward, and by about 40\% compared to no feedforward (PID only).

In summary, this work contributes: 
\begin{enumerate}
    \item the first complete kinematic and dynamic modeling formulation tailored to the dVRK-Si PSM;
    \item an excitation-trajectory-driven, physically consistent dynamic model parameter identification pipeline;
    \item two gravity compensation implementations, including a simplified variant with low computational requirements;
    \item comprehensive experimental evaluation demonstrating the improved control performance on the dVRK-Si platform;
    \item an open source implementation at \url{https://github.com/jhu-dvrk/dvrk_psm_dynamics_identification}.
\end{enumerate}

\begin{figure*}[ht]
    \centering
    \includegraphics[width=0.9\linewidth]{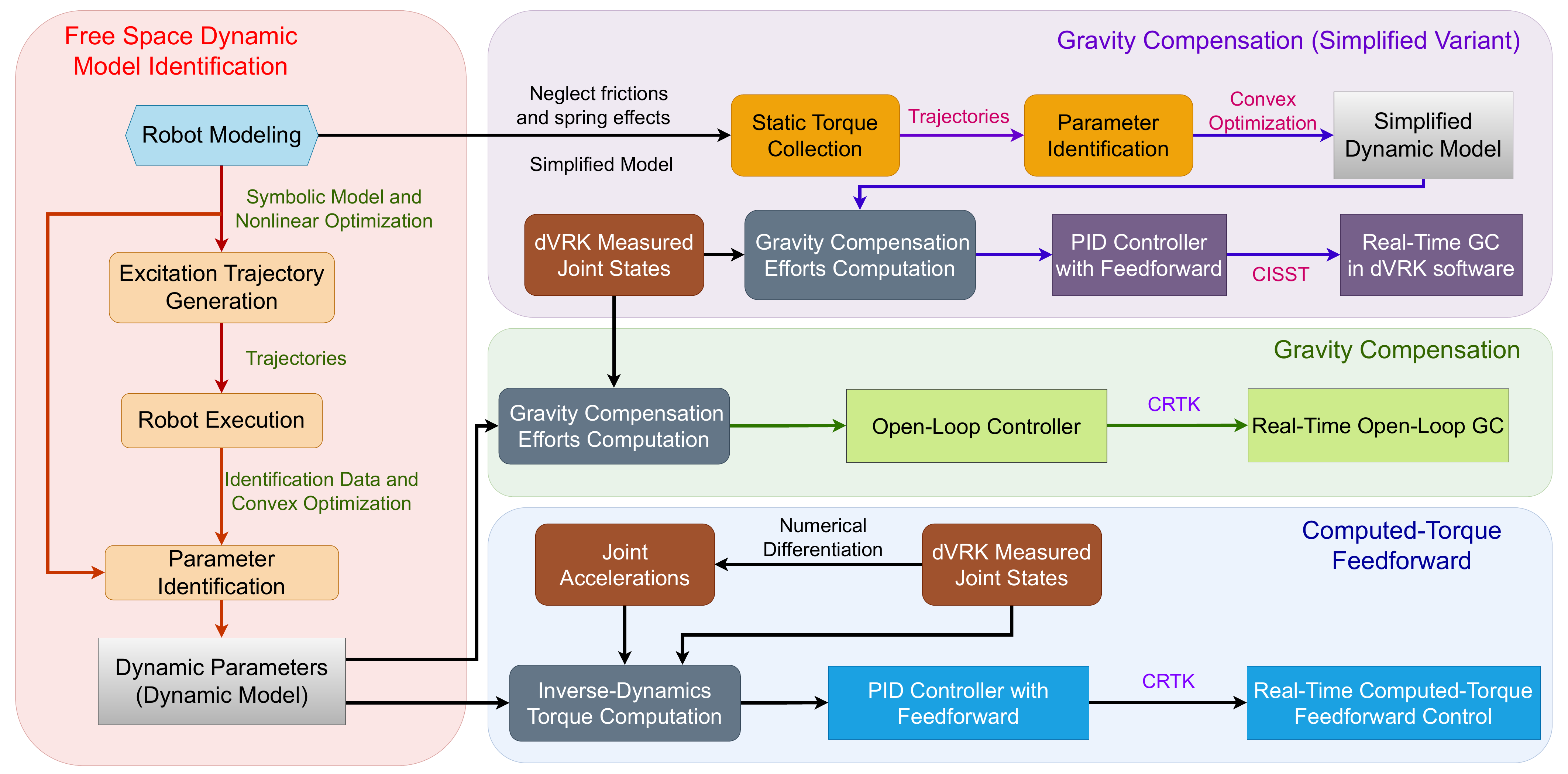}
    \caption{An overview of the workflow for the proposed dVRK-Si PSM modeling, dynamic parameter identification, and deployment. Free-space dynamic model identification proceeds from symbolic robot modeling to optimized periodic excitation-trajectory generation, hardware execution, and physically consistent parameter estimation via convex optimization to obtain the full dynamic model. The identified model is then deployed through three control implementations: (1) a simplified statics-only gravity-compensation variant fitted from static torque measurements (neglecting friction and spring effects) and integrated into the dVRK software, (2) real-time open-loop gravity compensation implementation, and (3) real-time computed-torque feedforward, where joint accelerations are obtained by numerical differentiation and inverse-dynamics torques are added as a feedforward term to a baseline PID loop.}
    \label{fig:overview}
\end{figure*}



\section{Related Work}

Over the past decade, substantial research efforts have advanced dynamic system identification for the dVRK Classic, with the goals of improving simulation fidelity~\cite{ferro2022coppeliasim}, control performance~\cite{lin2019reliable}, and force estimation~\cite{sang2017external, pique2019dynamic, yilmaz2020neural, wu2021robot, yang2024hybrid, yang2025effectiveness}. Fontanelli et al.~\cite{fontanelli2017modelling} identified the dynamic parameters of the Master Tool Manipulator (MTM) and PSM of the dVRK Classic using the method proposed in Sousa et al.~\cite{sousa2014physical}. Sang et al.~\cite{sang2017external} and Piqu{\'e} et al.~\cite{pique2019dynamic} estimated the dynamic parameters of the dVRK Classic PSM via Least-square Regression and leveraged the resulting models for sensorless external force estimation for force feedback. Wang et al.~\cite{wang2019convex} developed the first open-source framework for the dVRK Classic MTM  and PSM, addressing the prerequisite for deploying advanced model-based control algorithms, such as those in Lin et al.~\cite{lin2019reliable}. More recently, Argin et al. extended this framework by incorporating an improved friction model and employing an Augmented Lagrangian Particle Swarm Algorithm approach~\cite{argin2024davinci}. Nevertheless, all aforementioned PSM identification studies focus exclusively on the dVRK Classic and do not address the dVRK-Si platform. 

In parallel, rapid progress in artificial intelligence and machine learning has motivated learning-based alternatives for addressing the canonical identification and force estimation problem~\cite{pillonetto2022regularized}. Yilmaz et al. proposed a neural-network-based approach for external force estimation~\cite{yilmaz2020neural}. Zhang et al. extended this line of work to estimate both the external wrench and grasping force~\cite{zhang2022learning}, and Wu et al. further introduced a correction network to compensate for trocar–instrument interactions on an abdominal phantom~\cite{wu2021robot}. Subsequent work has leveraged transfer learning to better accommodate pseudo-clinical settings, enabling force estimation to generalize across different robots, instruments, and ports on an abdominal phantom~\cite{yilmaz2022transfer}. Despite their promise, learning-based approaches remain highly sensitive to the quality and coverage of training data; in particular, a mismatch in the training and test workspaces can result in high estimation errors~\cite{yang2024hybrid}. To mitigate this limitation, Yang et al.~\cite{yang2024hybrid} proposed a hybrid framework that combines the model-based and learning-based approaches and demonstrated improved force estimation accuracy on the dVRK Classic PSM. Preliminary learning-based investigations on dVRK-Si PSM force estimation and calibration have also been reported~\cite{yang2025effectiveness, chen2025accuracy}, although their reliability may be limited by persistent concerns regarding robustness and generalization.

To the best of our knowledge, the only publicly reported model-based dynamic identification study for the dVRK-Si PSM is our preliminary version presented at the Hamlyn Symposium on Medical Robotics~\cite{zhou2025gravity}. The present manuscript substantially extends~\cite{zhou2025gravity} with complete derivations, expanded methodology details, and extensive experimental validation.

\section{Methodology}

The equations of motion for a robotic manipulator can be expressed using the well-known relation:
\begin{equation}
    \begin{split}
         \tau = M(q) \Ddot{q} + C(q, \Dot{q}) + G(q)
    \end{split}
    \label{eq:robotdyn}
\end{equation}
where $q$ represents the joint variables, $\tau$ represents the joint torques, $M(q)$ represents the mass matrix of the robot, and $C(q,\Dot{q})$, $G(q)$ represent the Coriolis and gravity vectors.

Our proposed approach is as follows (Fig.~\ref{fig:overview}):
\begin{enumerate}[label=(\alph*)]
    \item Build the kinematic model of the dVRK-Si PSM based on the parameters in Table~\ref{tab:psm_para};
    \item Construct the dynamic model of the PSM using the Euler-Lagrangian approach~\cite{wang2019convex};
    \item Calculate the optimal excitation trajectory based on the given joint constraints and the dynamic model;
    \item Run the excitation trajectory on the physical dVRK-Si PSM to collect kinematic and dynamic data, and perform preprocessing;
    \item Solve for the model parameters using a convex optimization approach;
    \item Evaluate the model at runtime to enable gravity compensation (using only the $G(q)$ term) or computed-torque feedforward (all terms).
\end{enumerate}

Our work builds upon the framework of Wang et al.~\cite{wang2019convex} with an improved Coulomb friction model and excitation trajectory generation approach and delivers a Python 3-compatible, open-source package for dynamic model identification. In addition, we integrate the simplified gravity compensation variant into the dVRK C++ software stack to enable real-time (\textgreater 1\,kHz) execution.
%

\subsection{Kinematic Model}

\subsubsection{Joint Coordinates}

\begin{figure}[ht]
    \centering
    \includegraphics[width=\linewidth]{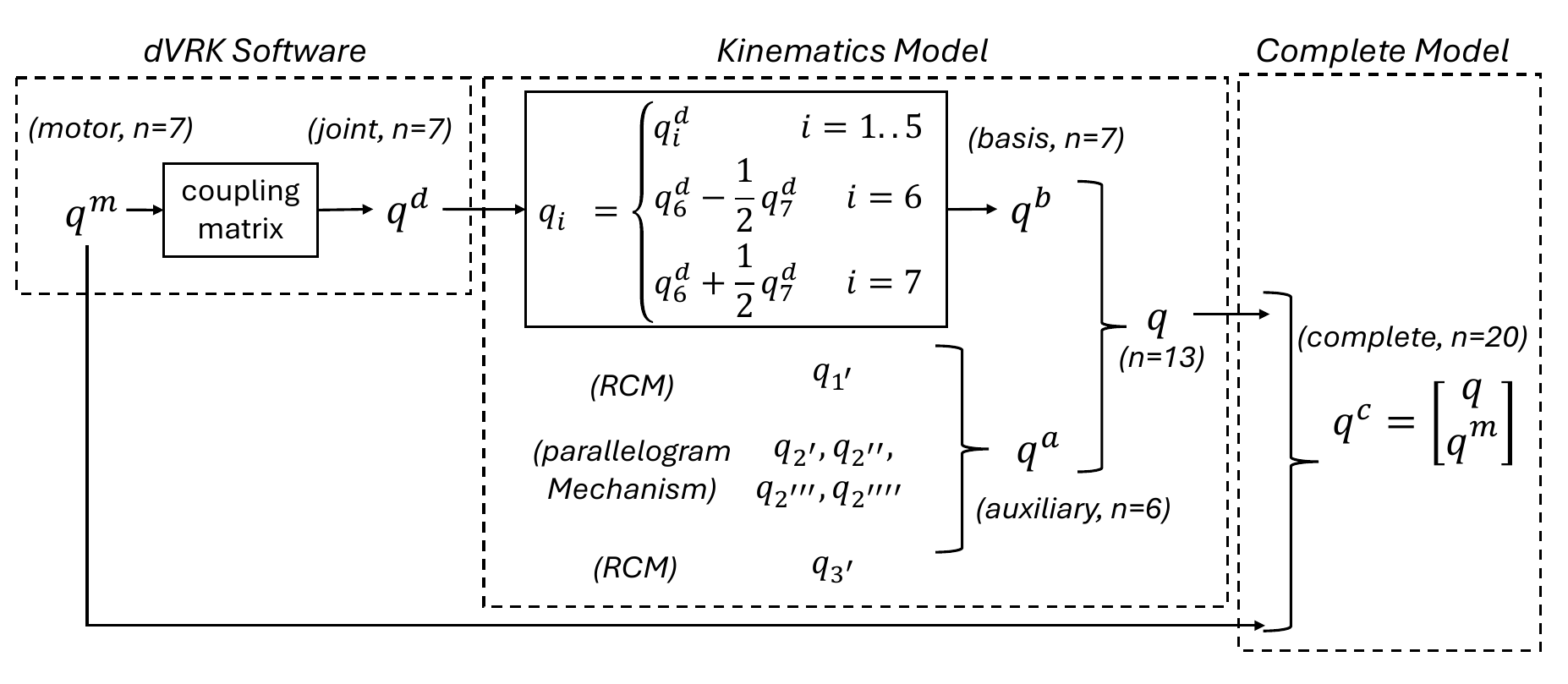}
    \caption{Overview of the joint coordinates provided by dVRK software (n=7), extended for dynamics estimation (n=13), and for complete model (n=20).}
    \label{fig:joint_coords}
\end{figure}

To build the relationship between the robot joint motion in the dVRK package~\cite{kazanzides2014open} and the torque of each motor, multiple types of joint coordinates are defined (Fig.~\ref{fig:joint_coords}):

\begin{itemize}
    \item $\bm{q}^d$: the joint coordinates used in the dVRK package on the dVRK-Si PSM. These are related to the motor coordinates $\bm{q}^m$ via a coupling matrix.
    \item $\bm{q}^b$: the basis joint coordinates which can adequately and independently represent the kinematics of the dVRK-Si PSM. Since the PSM has seven actuated degrees of freedom (DOF), the basis joint coordinates can be represented by $\bm{q}^b = \begin{bmatrix} q_1 & q_2 & \cdots & q_7 \end{bmatrix}^T$.
    \item $\bm{q}^a$: the six auxiliary joint coordinates required to model the RCM or the parallelogram mechanism, which can be represented by a linear combination of $\bm{q^b}$ (for example, $q_{2''} = - q_2$).
    \item $\bm{q}$: the joint coordinates used in the kinematic modeling of the dVRK-Si PSM in this work (the first 13 parameters in Table~\ref{tab:psm_para}), represented by $\bm{q} = \begin{bmatrix}  (\bm{q}^b)^T & (\bm{q}^a)^T \end{bmatrix}^T$.
    \item $\bm{q}^m$: the motor coordinates (called \emph{actuator space} in the dVRK software), which are used to model motor inertia and transmission effects. 
    \item $\bm{q}^c$: the complete joint coordinates used to attach friction, motor inertia and elastic elements, represented by $\bm{q^c} = \begin{bmatrix}  \bm{q}^T & (\bm{q}^m)^T \end{bmatrix}^T$.
\end{itemize}



\subsubsection{Joint Coordinate Mapping}

We model the first five joints of the PSM identical to the dVRK software package, which means $q_{1-5} = q^b_{1-5} = q^d_{1-5}$. 
The dVRK package models the last two joints as follows:
\begin{itemize}
    \item $q^d_6$: the angle from the insertion axis to the bisector of the two jaw tips.
    \item $q^d_7$: the angle between the two jaw tips.
\end{itemize}
For convenience, we henceforth use $q$, rather than $q^b$, when referring to any of the first seven joints.

\begin{figure}[ht]
    \centering
    \includegraphics[width=0.8\linewidth]{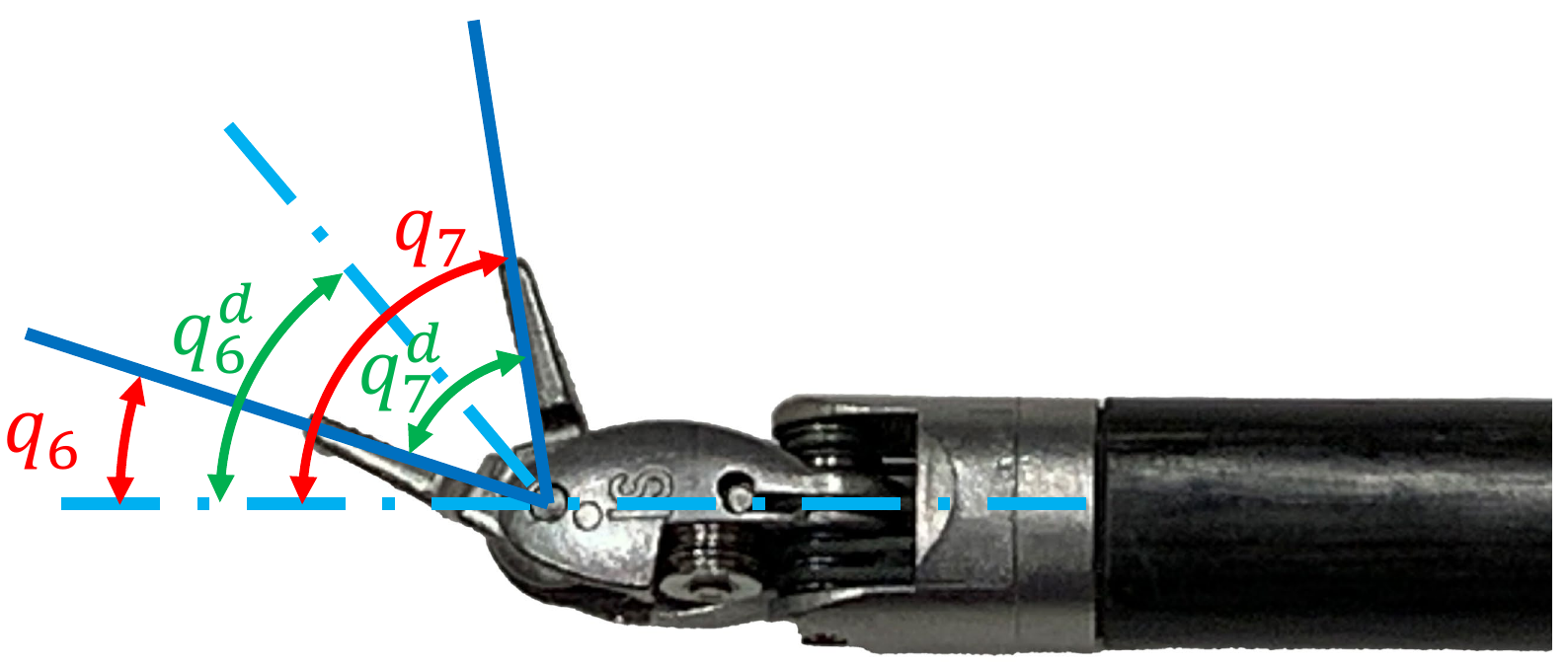}
    \caption{Modeling of the motion of the PSM gripper.}
    \label{fig:gripper_motion}
\end{figure}

As shown in Fig.~\ref{fig:gripper_motion}, the relationship between $q_6$, $q_7$ and $q^d_6$, $q^d_7$ can be represented by Eq.~\ref{eq:gripper_trans}.
\begin{equation}
   \begin{split}
       \begin{bmatrix} q^d_6 & q^d_7 \end{bmatrix} = \begin{bmatrix} \frac{q_6 + q_7}{2} & -q_6+q_7 \end{bmatrix}
   \end{split}
   \label{eq:gripper_trans}
\end{equation}

Meanwhile, since the first four joints are independently driven, the motor coordinates are identical to the joints, which can be represented by $q^d_{1-4} = q^m_{1-4}$. The coupling matrix $A_m^d$, which maps $q^m_{5-7}$ to $q^d_{5-7}$, is specific to each instrument and is provided by the dVRK software package~\cite{kazanzides2014open}; for the Large Needle Driver it is:
{\footnotesize
\begin{equation}
    \begin{split}
        \begin{bmatrix} q^d_5 \\ q^d_6 \\ q^d_7 \end{bmatrix} = A_m^d \begin{bmatrix} q^m_5 \\ q^m_6 \\ q^m_7 \end{bmatrix} = \begin{bmatrix} 1.0186 & 0 & 0 \\ -0.8306 & 0.6089 & 0.6089 \\ 0 & -1.2177 & 1.2177 \end{bmatrix} \begin{bmatrix} q^m_5 \\ q^m_6 \\ q^m_7 \end{bmatrix}
    \end{split}
    \label{eq:motor_couple}
\end{equation}
}
\subsubsection{Kinematic Modeling of the dVRK-Si PSM}

\begin{figure}[ht]
    \centering
    \includegraphics[width=\linewidth]{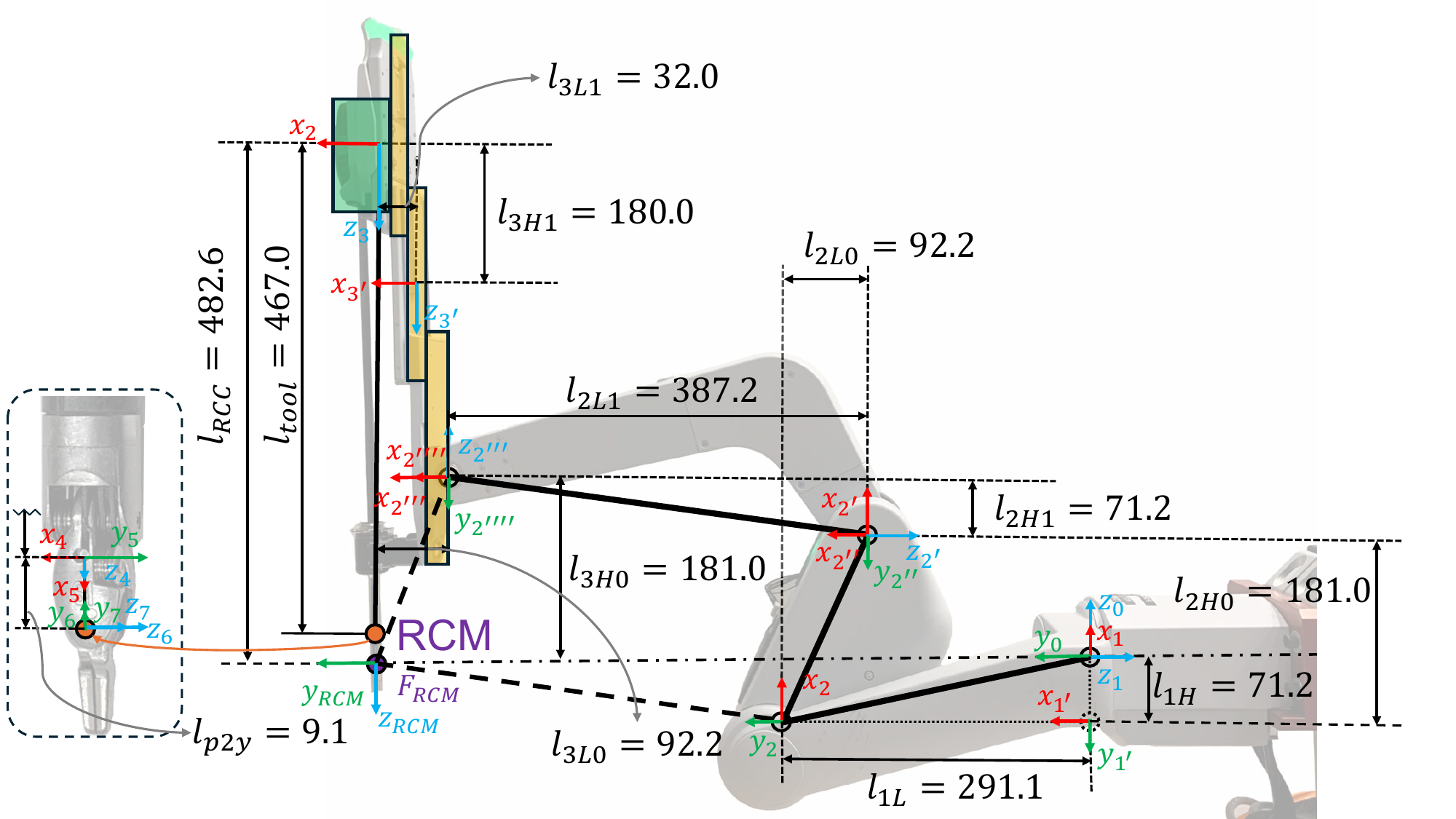}
    \caption{The planar view of the frame definition for the dVRK-Si PSM. The units of the link lengths are mm. All frames are left-hand frames.}
    \label{fig:kinematics}
\end{figure}

Based on our own measurements and manufacturer-provided specifications, we formulate the complete kinematic model of the dVRK-Si PSM, as shown in Fig~\ref{fig:kinematics}. The detailed kinematic model parameters are summarized in Table~\ref{tab:psm_para}.




\begin{table}[ht]
\caption{dVRK-Si PSM Model Description Parameters}
\resizebox{\columnwidth}{!}{%
\begin{tabular}{llllllllll}
$i$      & Ref   & $a_{i-1}$ & $\alpha_{i-1}$   & $d_i$                    & $\theta_i$               & $\delta_{Li}$ & $I_{mi}$     & $F_i$        & $K_{si}$       \\
\hline
1        & 0     & 0         & $\frac{\pi}{2}$  & 0                        & $q_1+\frac{\pi}{2}$      & $\checkmark$  & $\times$     & $\checkmark$ & $\times$     \\
1'       & 1     & $-l_{1H}$ & $-\frac{\pi}{2}$ & 0                        & $\frac{\pi}{2}$          & $\times$      & $\times$     & $\times$     & $\times$     \\
2        & 1'    & $l_{1L}$  & 0                & 0                        & $q_2-\frac{\pi}{2}$      & $\checkmark$  & $\times$     & $\checkmark$ & $\times$     \\
2'       & 2     & $l_{2L0}$ & $\frac{\pi}{2}$  & $l_{2H0}$                & 0                        & $\times$      & $\times$     & $\times$     & $\times$     \\
2''      & 2'    & 0         & $-\frac{\pi}{2}$ & 0                        & $- q_2 + \frac{\pi}{2}$  & $\checkmark$  & $\times$     & $\checkmark$ & $\times$     \\
2'''     & 2''   & $l_{2L1}$ & $\frac{\pi}{2}$  & $l_{2H1}$                & 0                        & $\times$      & $\times$     & $\times$     & $\times$     \\
2''''    & 2'''  & 0         & $-\frac{\pi}{2}$ & 0                        & $q_2$                    & $\checkmark$  & $\times$     & $\checkmark$ & $\times$     \\
3        & 2'''' & $l_{2L2}$ & $-\frac{\pi}{2}$ & $q_3+l_{c2}$  & 0                        & $\checkmark$  & $\times$     & $\checkmark$ & $\times$     \\
3'       & 3     & $-l_{3L}$ & 0                & $l_{3H} - \frac{q_3}{2}$ & 0                        & $\times$      & $\times$     & $\checkmark$ & $\times$     \\
4        & 3     & 0         & 0                & $l_{tool}$               & $q_4$                    & $\times$      & $\checkmark$ & $\checkmark$ & $\checkmark$ \\
5        & 4     & 0         & $\frac{\pi}{2}$  & 0                        & $q_{5}+ \frac{\pi}{2}$ & $\times$      & $\checkmark$ & $\checkmark$ & $\times$     \\
6        & 5     & $l_{p2y}$ & $-\frac{\pi}{2}$ & 0                        & $q_{6}+ \frac{\pi}{2}$ & $\times$      & $\times$     & $\checkmark$ & $\times$     \\
7        & 5     & $l_{p2y}$ & $-\frac{\pi}{2}$ & 0                        & $q_{7}+ \frac{\pi}{2}$ & $\times$      & $\times$     & $\checkmark$ & $\times$     \\
$^*F_{67}$ & -     & 0         & 0                & 0                        & $q_{7}-q_{6}$        & $\times$      & $\times$     & $\checkmark$ & $\times$     \\
$^*M_6$    & -     & 0         & 0                & 0                        & $q^m_6$                    & $\times$      & $\checkmark$ & $\checkmark$ & $\times$     \\
$^*M_7$    & -     & 0         & 0                & 0                        & $q^m_7$                    & $\times$      & $\checkmark$ & $\checkmark$ & $\times$       
\end{tabular}%
}
\footnotesize{Notes for the table: 
(1) Ref stands for the reference frame of link $i$.
(2) $a_{i-1}$, $\alpha_{i-1}$, $d_i$ and $\theta_i$ are the modified DH parameters of link $i$. 
(3) $\delta_{Li}$, $I_{mi}$, $F_i$ and $K_{si}$ are the parameters of link inertia, motor inertia, joint friction, and torsional spring stiffness of link $i$, respectively. 
(4) $l_{c2}=l_{3H0} - l_{RCC}$.}
(5) $^*$ denotes that the parameters are not for the  kinematic model description. $M_6$ and $M_7$ represent the motor model of joint 6 and 7. $F_{67}$ represents the relative motion model between link 6 and 7. $q^m_i$ represents the equivalent motor movements at joint $i$. 
\label{tab:psm_para}
\end{table}

\subsection{Dynamic Parameters}

For each link $k$, its dynamic parameter set comprises two components:
\begin{itemize}
    \item the parameters associated with the link's mass and inertia properties;
    \item additional parameters capturing internal effects, such as frictions, motor inertia and spring stiffness.
\end{itemize}

\subsubsection{Link Inertial Parameters}

Each rigid link $k$ is characterized by its mass $m_k$, the position of its center of mass (COM) $\bm{r}_k$ with respect to link frame $k$, and its inertia tensor $\bm{I}_{C_k}$ about the COM. To express the equations of motion in a form that is linear in the dynamic parameters, we adopt the barycentric parameterization~\cite{raucent1992identification}, where mass $m_k$, the first moment of mass $\bm{h}_k \in \mathbb{R}^3 = m_k \bm{r}_k$ and inertia tensor $\bm{I}_k$ about frame $k$ are used to form the dynamic parameter vector. The inertia tensor $\bm{I}_k$ can be calculated using the parallel axis theorem:
{\footnotesize
\begin{equation}
    \begin{split}
        \bm{I}_k = \bm{I}_{C_k} + m_k \bm{S}\left ( \frac{\bm{h}_k}{m_k} \right )^T \bm{S} \left ( \frac{\bm{h}_k}{m_k} \right ) = \begin{bmatrix} I_{kxx} & I_{kxy} & I_{kxz} \\ I_{kxy} & I_{kyy} & I_{kyz} \\ I_{kxz} & I_{kyz} & I_{kzz} \end{bmatrix}
    \end{split}
    \label{eq:inertia_frame}
\end{equation}
}
where $\bm{S}(\cdot)$ denotes the skew-symmetric operator.

The inertial parameters of link $k$ can then be concatenated into a single parameter vector $\bm{\theta}_{Lk} \in \mathbb{R}^{10}$ as:
{\footnotesize
\begin{equation}
    \begin{split}
        \bm{\theta}_{Lk} =  \begin{bmatrix} I_{kxx} & I_{kxy} & I_{kxz} & I_{kyy} & I_{kyz} & I_{kzz} & \bm{h}^T_k & m_k \end{bmatrix}^T
    \end{split}
    \label{eq:param_link}
\end{equation}
}

\subsubsection{Additional Parameters}

In addition to the inertial parameters of link $k$, we include the associated joint friction coefficients, motor inertia $I_{mk}$, and torsional spring stiffness $K_{sk}$ as an additional set of parameters:
\begin{equation}
    \begin{split}
        \bm{\theta}_{Ak} =  \begin{bmatrix} F_{vk} & F_{ck} & F_{bk} & I_{mk}  & K_{sk} \end{bmatrix}^T
    \end{split}
    \label{eq:param_other}
\end{equation}
where $F_{vk}$, $F_{ck}$ denote the viscous and Coulomb friction coefficients, $F_{bk}$ denotes a constant friction bias, and $I_{mk}$ denotes the equivalent motor inertia reflected to joint $k$ through the transmission. Unless otherwise specified, we adopt the indexing convention that link $k$ is actuated by joint $k$.

Eventually, the parameters associated with all $n$ joints are concatenated to form the full dynamic parameters $\bm{\theta}$ of the dVRK-Si PSM:
\begin{equation}
    \begin{split}
        \bm{\theta} = \begin{bmatrix} \bm{\theta}^T_{L1} & \bm{\theta}^T_{A1} & \cdots & \bm{\theta}^T_{Ln} & \bm{\theta}^T_{An} \end{bmatrix}^T
    \end{split}
    \label{eq:param_all}
\end{equation}
where $n=16$ according to the model description parameters summarized in Table~\ref{tab:psm_para}.

\subsection{Inverse Dynamics Formulation}

The inverse dynamic model for the robot with parallelogram mechanism can be calculated using Newton-Euler or Euler-Lagrange methods~\cite{sciavicco2012modelling} for the equivalent tree structure and by considering kinematic constraints between joint coordinates~\cite{murray1989dynamic}. In this work, we select the Euler-Lagrange approach to model the dynamics of the dVRK-Si PSM since it can handle the kinematic constraints more intuitively. 

\subsubsection{Link Inertia Torque}

The Lagrangian $L$ is calculated as the difference of the kinetic energy $K$ and potential energy $P$ of the robot, $L = K - P$. Friction, motor inertia, and spring stiffness are not included in $L$ and are modeled separately.

The relationship between motor motion $\bm{q}^m$ and the torque of each motor $i$ caused by link inertia can be represented by:
\begin{equation}
    \begin{split}
        \tau^m_{LIi}(\bm{q}^m) = \frac{d}{dt} \frac{\partial L}{\partial \Dot{q}^m_i} - \frac{\partial L}{\partial q^m_i} 
    \end{split}
    \label{eq:tq_link}
\end{equation}

\subsubsection{Motor Inertia Torque}

The torques caused by motor inertia can be represented by:
\begin{equation}
    \begin{split}
        \bm{\tau}^m_{MI}(\bm{\Ddot{q}}^m) = \bm{I}_m \bm{\Ddot{q}}^m
    \end{split}
    \label{eq:tq_motor}
\end{equation}
where $\bm{I}_m$ is a diagonal matrix of the motor inertia coefficients of all motors.

\subsubsection{Joint Friction Torque}

The torques induced by friction on all joints $\bm{q}^c$ can be represented by:
\begin{equation}
    \begin{split}
        \bm{\tau}^c_f(\bm{\Dot{q}}^c) = \bm{F}_v \bm{\Dot{q}}^c + \bm{F}_c \tanh{\left ( \bm{\epsilon} \bm{\Dot{q}}^c \right )} + \bm{F}_b
    \end{split}
    \label{eq:tq_fricition}
\end{equation}
where $\bm{F}_v$ and $\bm{F}_c$ are diagonal matrices of the viscous and Coulomb friction coefficients, $\bm{F}_b$ denotes the vector of constant friction bias across all joints, and $\bm{\epsilon}$ denotes the diagonal matrix of the joint-velocity-smoothing parameters which can provide a sufficiently wide transition band for near-zero velocities to mitigate sign-switching caused by velocity estimation noise, while remaining close to the signum function at typical operating velocities.


Notably, we utilize the $\tanh{(\cdot)}$ in place of the discontinuous signum function~\cite{green2012modelling} to obtain a continuous and differentiable approximation of Coulomb friction, thereby improving numerical robustness and reducing sensitivity to velocity noise in the vicinity of $\bm{\Dot{q}}^c=\bm{0}$.

\subsubsection{Spring Stiffness Torque}

The torques due to the spring stiffness can be represented by
\begin{equation}
    \begin{split}
        \bm{\tau}^c_s(\bm{q}^c) = \bm{K}_s \bm{\Delta l}_s
    \end{split}
    \label{eq:tq_spring}
\end{equation}
where $\bm{K}_s$ is a diagonal matrix of the spring stiffness constants of all springs, and $\bm{\Delta l}_s$ denotes the equivalent prolongation vector which can be represented by a linear combination of $\bm{q}^c$.

\subsubsection{Project non-link torques to motor space}

The joint torques caused by springs and frictions can be projected onto the motor joints, using the Jacobian matrix of their corresponding complete joint coordinates $\bm{q}^c$ with respect to the motor joint coordinates $\bm{q}^m$~\cite{murray1989dynamic}. Therefore, the overall motor torques $\bm{\tau}^m$ can be represented by
{\footnotesize
\begin{equation}
    \begin{split}
        \bm{\tau}^m(\bm{q}^m, \bm{\Dot{q}}^m, \bm{\Ddot{q}}^m) &= \bm{\tau}^m_{LI}(\bm{q}^m) + \bm{\tau}^m_{MI}(\bm{\Ddot{q}}^m) + \frac{\partial \bm{q}^c}{\partial \bm{q}^m}[\bm{\tau}^c_s(\bm{q}^c)+\bm{\tau}^c_f(\bm{\Dot{q}}^c)] \\
        &=\hat{M}(\bm{q}^m) \bm{\Ddot{q}}^m + \hat{C}(\bm{q}^m, \bm{\Dot{q}}^m) + \hat{G}(\bm{q}^m)
    \end{split}
    \label{eq:tq_all}
\end{equation}
}
where $\bm{\tau}^m_{LI}(\cdot)$ denotes the link inertial contribution, $\bm{\tau}^m_{MI}(\cdot)$ denotes the motor inertial contribution, and $\bm{\tau}^c_s(\cdot)$ and $\bm{\tau}^c_f(\cdot)$ represent the spring-induced and friction-induced torques defined in the complete joint-coordinate space, respectively. Henceforth, we use $\hat{M}$, $\hat{C}$, and $\hat{G}$ to denote the augmented mass matrix, augmented Coriolis/velocity-dependent vector, and augmented gravity/load-dependent vector, respectively, after incorporating the effects of motor inertia, elastic elements, and joint friction into the motor-joint dynamics.

\subsection{Linear Parameterization}

\subsubsection{Linear Regressor Formulation}

To identify the dynamic parameters $\bm{\theta}$ in Eq.~\ref{eq:param_all}, Eq.~\ref{eq:tq_all} can be rewritten linearly in $\bm{\theta}$~\cite{maes1989linearity}:
{\small
\begin{equation}
    \begin{split}
        \bm{\tau}^m(\bm{q}^m, \bm{\Dot{q}}^m, \bm{\Ddot{q}}^m) &=\hat{M}(\bm{q}^m) \bm{\Ddot{q}}^m + \hat{C}(\bm{q}^m, \bm{\Dot{q}}^m) + \hat{G}(\bm{q}^m) \\
        &= \bm{Y}(\bm{q}^m, \bm{\Dot{q}}^m, \bm{\Ddot{q}}^m) \bm{\theta}
    \end{split}
    \label{eq:reg_init}
\end{equation}
}
where $\bm{Y}(\cdot)$ is the inverse dynamics regressor matrix, which is also known as the regressor.

\subsubsection{Base Parameters Formulation}

Because of the structural dependence (the parallelogram mechanism), the dynamics depends only on certain linear combinations of the parameters, which are the base parameters. It is a minimum set of dynamics parameters which can fully describe the dynamic model of the robot. 

To find the base parameters, we introduce a permutation matrix $P \in \mathbb{R}^{p \times p}$, where the regressor $\bm{Y} \in \mathbb{R}^{n \times p}$ and $n$ denotes the number of joints, as shown in Eq.~\ref{eq:param_all}. Then, Eq.~\ref{eq:reg_init} can be written as:
\begin{equation}
    \begin{split}
        \bm{Y}P &= \begin{bmatrix} \bm{Y}_b & \bm{Y}_d \end{bmatrix}, \quad P^T \bm{\theta} = \begin{bmatrix} \bm{\theta}_b \\ \bm{\theta}_d \end{bmatrix} \\
        \bm{\tau}^m &= \bm{Y} \bm{\theta} = \begin{bmatrix} \bm{Y}_b & \bm{Y}_d \end{bmatrix} \begin{bmatrix} \bm{\theta}_b \\ \bm{\theta}_d \end{bmatrix}
    \end{split}
    \label{eq:Y_decomp}
\end{equation}
where $\bm{Y}_b \in \mathbb{R}^{n \times n_b}$ has $n_b$ linearly independent columns, $\bm{Y}_d \in \mathbb{R}^{n \times n_d}$ contains the remaining dependent or null column, and $p = n_b + n_d$. In this study, the decomposition yields $n_b = 48$ identifiable base parameters and $n_d = 126$ dependent parameters.

Since $\bm{Y}_d$ can be written as a linear combination of $\bm{Y}_b$, there exists a constant matrix $B_d \in \mathbb{R}^{n_b \times n_d}$ such that
\begin{equation}
    \begin{split}
        \bm{Y}_d = \bm{Y}_b B_d
    \end{split}
    \label{eq:Y_bd}
\end{equation}

Substituting Eq.~\ref{eq:Y_bd} into Eq.~\ref{eq:Y_decomp}, we can obtain:
\begin{equation}
    \begin{split}
        \bm{\tau}^m &= \bm{Y}_b \bm{\beta} \\
        \bm{\beta} &= \bm{\theta}_b + B_d \bm{\theta}_d
    \end{split}
    \label{eq:param_base}
\end{equation}
where $\bm{\beta}$ denotes the base parameter vector and $\bm{Y}_b$ is the regressor for the base parameters.

In our work, we utilize the QR decomposition with pivoting~\cite{gautier1991numerical} to numerically find the base parameters from Eq.~\ref{eq:reg_init}.

\subsection{Excitation Trajectory Design}

\subsubsection{Problem Formulation}

To find the periodic excitation trajectories, we utilize a Fourier series to generate data for dynamic model identification~\cite{swevers2002optimal}. The joint coordinate $q^m_j$ of motor $j$ can be calculated by:
{\small
\begin{equation}
    \begin{split}
        q_j^m(t; \bm{x}) = q_{oj}^m + \sum_{k=1}^{n_H}\left [ \frac{a_{jk}}{k\omega_f}\sin{(k\omega_f t)} + \frac{b_{jk}}{k\omega_f}\cos{(k\omega_f t)} \right ]
    \end{split}
    \label{eq:gen_joint}
\end{equation}
}
where $\omega_f = 2 \pi f_f$ is the angular component of the fundamental frequency $f_f$, $n_H$ is the harmonic number of the Fourier series, $a_{jk}$ and $b_{jk}$ are the amplitudes of the $j^{th}$-order sine and cosine function, $q_{oj}^m$ is the position offset, $t$ is time, and $\bm{x}$ denotes the coefficients to be optimized, as shown in Eq.~\ref{eq:traj_gen_opt_vec}.
{\small
\begin{equation}
    \begin{split}
        \bm{x} &= \begin{bmatrix} q_{o1}, a_{11}, b_{11},\cdots,a_{1n_H},\cdots,q_{on_s},\cdots, a_{n_sn_H} ,b_{n_sn_H}\end{bmatrix}
    \end{split}
    \label{eq:traj_gen_opt_vec}
\end{equation}
}
$n_s$ in Eq.~\ref{eq:traj_gen_opt_vec} denotes the number of sampling points in one period $T_s = \frac{1}{f_f}$.

Given the joint coordinates $q^m_j(t)$ shown in Eq.~\ref{eq:gen_joint}, we can calculate the motor joint velocity $\Dot{q}_j^m(t)$ and acceleration $\Ddot{q}_j^m(t)$ using the first and second derivative of $q^m_j(t)$, as shown in Eq.~\ref{eq:gen_vel_acc}.
{\small
\begin{equation}
    \begin{split}
        \Dot{q}_j^m(t; \bm{x}) &=  \sum_{k=1}^{n_H} [a_{jk}\cos{(k\omega_f t)} - b_{jk}\sin{(k\omega_f t)}] \\
        \Ddot{q}_j^m(t; \bm{x}) &= - \sum_{k=1}^{n_H} (k\omega_f) [a_{jk}\sin{(k\omega_f t)} + b_{jk}\cos{(k\omega_f t)}]
    \end{split}
    \label{eq:gen_vel_acc}
\end{equation}
}
Substituting the calculated $q^m_j(t)$, $\Dot{q}_j^m(t)$ and $\Ddot{q}_j^m(t)$ into Eq.~\ref{eq:param_base}, we can stack all samples into one data matrix:
\begin{equation}
    \begin{split}
        \begin{bmatrix} \tau^m(t_1) \\ \tau^m(t_2) \\ \vdots \\ \tau^m(t_{n_s})\end{bmatrix} = \begin{bmatrix} Y_b(t_1) \\ Y_b(t_2) \\ \vdots \\ Y_b(t_{n_s})\end{bmatrix} \bm{\beta} = \bm{W}_b(\bm{x}) \bm{\beta}
    \end{split}
    \label{eq:form_wb}
\end{equation}
where $\bm{W}_b(\bm{x})$ denotes the regression matrix for the excitation trajectory generation problem.

Therefore, the excitation trajectory design problem amounts to choosing $\bm{x}$ to yield a well-conditioned $\bm{W}_b(\bm{x})$, which can improve the numerical stability and accuracy of the parameter identification.

\subsubsection{Nonlinear Optimization}

To obtain a well-conditioned $\bm{W}_b(\bm{x})$, we minimize the L2-norm condition number $\kappa(\cdot)$ of the matrix $\bm{W}_b(\bm{x})$:
\begin{equation}
    \begin{split}
        \kappa(\bm{W}_b) = \frac{\sigma_{max}(\bm{W}_b)}{\sigma_{min}(\bm{W}_b)}
    \end{split}
    \label{eq:cond_init}
\end{equation}
where $\sigma(\cdot)$ denotes the singular value of the matrix. 



Then, we scale each column of $\bm{W}_b$ by dividing each element by the column's Root Mean Square (RMS), and obtain the RMS-scaled matrix $\bm{\Tilde{W}}_b$. We utilize the RMS-scaled matrix for optimization to avoid scale domination~\cite{ruiz2001scaling}.

The optimization is constrained by the given joint position and velocity limits. Moreover, we optimize the log-condition number $\log{(\kappa(\cdot))}$ since the logarithm compresses the wide dynamic range of $\kappa(\cdot)$ and improves numerical stability and optimizer convergence by reducing sensitivity when $\kappa(\cdot)$ becomes large~\cite{chen2011minimizing}. Therefore, the overall optimization problem can be presented as:
\begin{equation}
    \begin{split}
        \min_{\bm{x}} \quad & J_{traj}(\bm{x}) = \log{(\kappa(\bm{\Tilde{W}}_{b}(\bm{x})))} \\
        \text{s.t.}\quad 
        & q_{j,\min}^m \leq q^m_j(t) \leq q_{j, \max}^m, \qquad \forall\, j \\
        & \dot{q}_{j, \min}^m \leq \dot{q}^m_j(t) \leq \dot{q}_{j, \max}^,, \qquad \forall\, j
    \end{split}
\end{equation}
where $j$ denotes the motor joint $j$.

In our work, we select the fundamental frequency $f_f$ = 0.18\,Hz and the harmonic number $n_H$ = 6 to generate the excitation trajectory~\cite{wang2019convex}. Eventually, the constrained nonlinear optimization problem is solved using pyOpt~\cite{perez2012pyopt} with the Sequential Least Squares Quadratic Programming (SLSQP) algorithm~\cite{kraft1988software}, yielding the optimal coefficient vector $\bm{x}$.

\subsection{Data Collection and Preprocessing}

We collect data by executing the generated excitation trajectory at 200\,Hz using the dVRK software framework on a dVRK-Si PSM equipped with a Large Needle Driver, as shown in Fig.~\ref{fig:exp_setup}. Joint positions, velocities and torques are recorded by subscribing to the dVRK-ROS topics. Joint accelerations are then obtained via numerical differentiation of the measured velocities. Finally, the signals are preprocessed using a sixth-order low-pass filter with cutoff frequency of 5.4\,Hz, which was experimentally validated in Wang et al.~\cite{wang2019convex}. The filter was applied in both the forward and reverse directions to achieve zero-phase delay.

\subsection{Physical Consistency Constraints}
\label{subsec:pcc}

To reduce overfitting, we utilize physical consistency constraints when solving for the dynamic parameters.

\subsubsection{Constraints for Dynamic Model Identification for Gravity Compensation}

When solving the dynamic parameters for the gravity compensation problem, we use the physical consistency constraints as follows:
\begin{enumerate}[label=(\alph*)]
    \item The mass of each link is positive, $m_k > 0$.
    \item The COM of each link $k$, $\bm{r}_k$ is inside its convex hull, $ \bm{h}_k - m_k\bm{r}_{lk} \geq 0$ and  $m_k\bm{r}_{uk} - \bm{h}_k \geq 0$, where $\bm{r}_{lk}$ and $\bm{r}_{uk}$ denotes the lower and upper bounds of $\bm{r}_k$, respectively~\cite{sousa2014physical}.
    \item The viscous and Coulomb friction coefficients for each joint $k$ are positive, i.e. $F_{vk} > 0$ and $F_{ck} > 0$.
    \item The stiffness of spring $k$ is positive, i.e. $K_{sk} > 0$.
    \item The inertia of motor $k$ is positive, i.e. $I_{mk} > 0$.
\end{enumerate}

\subsubsection{Constraints for General Dynamic Model Identification}

When solving the dynamic parameters for the full dynamic system identification problem, we retain all of the aforementioned constraints and further impose one additional constraint as follows:
\begin{enumerate}[label=(\alph*)]
    \item The inertia matrix of each link $k$ is positive definite~\cite{yoshida2000verification}, and its eigenvalues should follow the triangle inequality conditions~\cite{traversaro2016identification}.
\end{enumerate}

The constraints regarding the mass and inertia properties of link $k$ can be derived into an equivalent with linear matrix inequalities (LMIs)~\cite{sousa2014physical}:
\begin{equation}
    \begin{split}
        \bm{D}_k(\bm{\theta}_{Lk}) = \begin{bmatrix} \frac{1}{2}\text{tr}(\bm{I}_k) \bm{I}_{3 \times 3} - \bm{I}_k & \bm{h}_k \\ \bm{h}_k^T & m_k \end{bmatrix} > \bm{0}
    \end{split}
\end{equation}
where $\bm{I}_{3 \times 3} \in \mathbb{R}^{3 \times 3}$ denotes the identity matrix.

We add the experimental lower and upper bounds to $m_k$, $F_{vk}$, $F_{ck}$ and $K_{sk}$ into the open-source dynamic model identification framework. 

\subsection{Model Parameter Identification}

The data is collected along with sampling time defined in Eq.~\ref{eq:form_wb}, and thus we can obtain the regression matrix $\bm{W}$ and dependent variable vector $\bm{b}$:
\begin{equation}
    \begin{split}
        \bm{W} = \begin{bmatrix} Y(t_1) \\ Y(t_2) \\ ... \\ Y(t_{n_s}) \end{bmatrix}, \qquad \bm{b} = \begin{bmatrix} \tau^m(t_1) \\ \tau^m(t_2) \\ ... \\ \tau^m(t_{n_s}) \end{bmatrix} 
    \end{split}
\end{equation}

Therefore, the identification problem can be formulated into an optimization problem which minimizes the scaled squared residual errors $\bm{J}(\bm{\theta})$ with respect to the decision vector $\bm{\theta}$:
\begin{equation}
    \begin{split}
        \bm{J}(\bm{\theta}) = \bm{\omega} || \bm{W} \bm{\theta} - \bm{b} ||_2^2
    \end{split}
\end{equation}
where the scale factor $\omega_i = \frac{1}{\max{(\bm{\tau}_i^m)} - \min{(\bm{\tau}_i^m)}}$ for each motor $i$.

We use the CVXPY package~\cite{diamond2016cvxpy} with the splitting conic solver (SCS)~\cite{o2017scs} to solve the convex optimization identification problem, subject to the constraints identified in Section~\ref{subsec:pcc}.

\subsection{Implementations using the Identified Dynamic Model}

All implementations which utilize the identified dynamic model run at 100\,Hz, matching the default ROS publishing rate of the dVRK software.

\subsubsection{Gravity Compensation}
\label{subsubsec: I-1}

For Gravity Compensation, we assume that the joint velocities $\bm{\Dot{q}}^c$ and accelerations $\bm{\Ddot{q}}^c$ are zero. 
We then substitute the estimated parameter $\bm{\hat{\theta}}$ into Eq.~\ref{eq:reg_init} to obtain:
\begin{equation}
    \begin{split}
        \bm{\tau}_g = \hat{G}(\bm{q}^d)
    \end{split}
    \label{eq:gc_qd}
\end{equation}

Then, we substitute the real-time joint value measurements $\bm{q}^d_{meas}$ from the dVRK software into Eq.~\ref{eq:gc_qd} to compute the real-time gravity force $\bm{\tau}_g$ in joint space. The resulting torque command is sent to the physical dVRK via ROS, using the CRTK convention~\cite{su2020collaborative}, to enable real-time gravity compensation in free space. This scheme is implemented open loop and the built-in PID controller is disabled during gravity compensation testing.

\subsubsection{Computed-Torque Feedforward}
\label{subsubsec: I-2}

Compared to the gravity compensation implementation, computed-torque feedforward accounts for joint dynamic movements, and thus the joint velocities and accelerations are no longer assumed to be zero. Substituting the estimated parameter $\bm{\hat{\theta}}$ into Eq.~\ref{eq:reg_init} yields:
\begin{equation}
    \begin{split}
         \bm{\hat{\tau}}_c = \hat{M}(\bm{q}^d) \bm{\Ddot{q}}^d + \hat{C}(\bm{q}^d, \bm{\Dot{q}}^d) + \hat{G}(\bm{q}^d)
    \end{split}
    \label{eq:ff_all}
\end{equation}
where $\bm{\hat{\tau}}_c$ is also known as the computed torques of the robot.

Then, the computed torques can be integrated into the dVRK controller as a feedforward term, $\bm{\tau}_{ff}$:
{\small
\begin{equation}
    \begin{split}
        \bm{\tau}_o &= \bm{K}_p (\bm{q}_{meas}^d - \bm{q}^d_{des}) + \bm{K}_d (\bm{\dot{q}}_{meas}^d - \bm{\dot{q}}^d_{des}) + \bm{\tau}_{ff} \\
        \bm{\tau}_{ff} &= \hat{M}(\bm{q}_{meas}^d) \bm{\Ddot{q}}_{des}^d + \hat{C}(\bm{q}_{meas}^d, \bm{\Dot{q}}_{meas}^d) + \hat{G}(\bm{q}_{meas}^d)
    \end{split}
\end{equation}
}
where $\bm{q}_{meas}^d$ and $\bm{q}_{des}^d$ represent measured and desired joint values, the joint accelerations $\bm{\Ddot{q}}_{des}^d$ are calculated by numerically differentiating the desired joint velocities, and $\bm{\tau}_o$ denotes the torque control output.


\subsection{Simplified Variant for Gravity Compensation Implementation in dVRK Software}
\label{sec:dvrk-gravity}

We additionally created a simplified statics-only variant of the gravity compensation model, suitable for easy integration into the dVRK's kilohertz frequency control loops. This model was developed by assuming $\bm{\dot{q}}^c = \bm{\ddot{q}}^c = 0$, and neglecting friction and spring effects, so that Eq.~\ref{eq:tq_all} simplifies to $\bm{\tau}_{rg} = \hat{G}_{rg}(\bm{q}^d)$. We used the same barycentric dynamics parameters as in Eq.~\ref{eq:param_link}, however in the static case we need only $h_{k}$ and $m_k$. Finally, the last four joints and links are lightweight and friction-dominated, so our simplified model only retains the first three joints/links.

One of the benefits of simplifying the model so drastically is the significantly lower dimensionality of the state space. Rather than using excitation trajectories to collect data for parameter estimation, it is feasible to exhaustively explore the joint space of the first three joints and measure joint torques while holding the robot stationary at each position. We collected joint torque measurements at a total of 150 distinct positions, visiting each position multiple times from different directions to mitigate static friction hysteresis.

As in Eq.~\ref{eq:reg_init}, this simplified model admits a linear formulation
\begin{equation}
    \begin{split}
        \bm{\tau}_{rg}(\bm{q}^d) = \bm{Y}_{rg}(\bm{q}^d) \bm{\theta}_{rg}
    \end{split}
    \label{eq:model_rg}
\end{equation}
where $\bm{\theta}_{rg}$ denotes the reduced statics parameters for simplified gravity compensation and $\bm{Y}_{rg}$ denotes the corresponding reduced regressor. 

Given $N$ pairs of joint positions and observed motor torques $(\bm{q}_i^d, \bm{\tau}_i^m)$ where $i$ denotes the $i^{th}$ sample, we fit the model by minimizing the least-squares objective $J_{rg}$:
\begin{equation}
    \begin{split}
        J_{rg}(\bm{\theta}_{rg}) = \frac{1}{2}\sum_{i=1}^{N}\left\lVert\bm{Y}_{rg}(\bm{q}_i^d) \bm{\theta}_{rg} - \bm{\tau}_i^m\right\rVert_2^2
    \end{split}
    \label{eq:g_simp_cost}
\end{equation}
subject to similar physical consistency constraints as the full dynamics model: (1) link masses $m_k$ must be positive, and (2) link centers of mass $r_k$ are contained within the link's bounding box. We used BFGS-B~\cite{byrd1995limited} to perform this bound-constrained minimization.

For integration into the dVRK arm controller (which runs at 1-2\,kHz depending on the configuration), we use the efficient recursive Newton-Euler algorithm, implemented in the open-source \emph{cisst} libraries \cite{deguet2008cisst}. For every controller step, our gravity compensation model is evaluated, and the predicted joint torques are added as a feed-forward to the joint-level controller. Notably, this simplified approach can handle different PSM mounting angles without additional tuning via modifying the mounting angle dVRK configuration setting.

This implementation of simplified gravity compensation has been available to the community in the dVRK software stack since February 2025.

\section{Experiment Setup}

\begin{figure}[ht]
    \centering
    \includegraphics[width=0.75\linewidth]{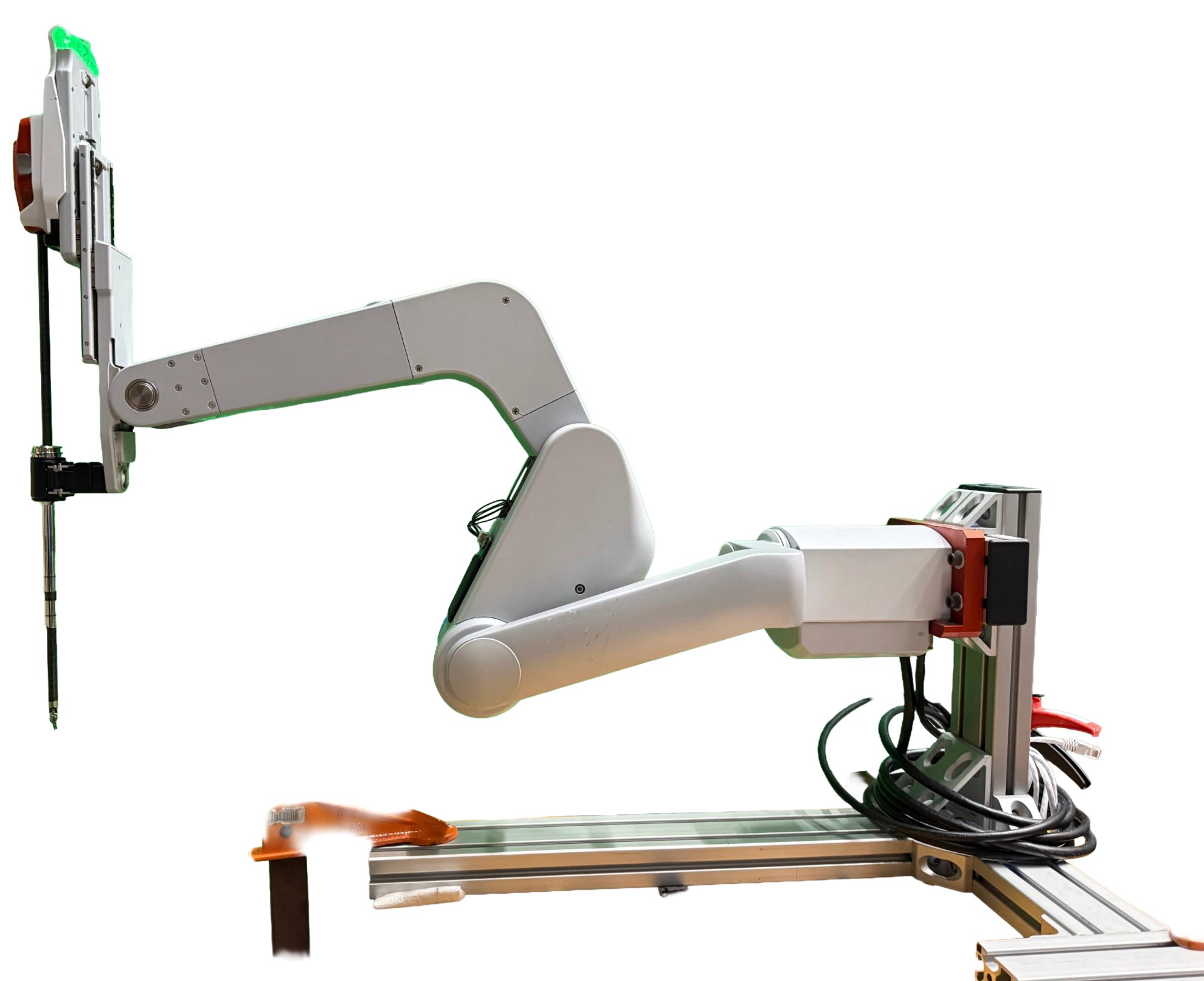}
    \caption{Experimental setup: dVRK-Si PSM is mounted at a zero-degree inclination relative to the table.}
    \label{fig:exp_setup}
\end{figure}

In our work, all control validation experiments are conducted on the dVRK-Si PSM setup shown in Fig.~\ref{fig:exp_setup}. The PSM is mounted at a zero-degree inclination relative to the table on an 80/20 aluminum frame and secured to the table using three clamps. Additional parameter identification accuracy evaluations were conducted on dVRK-Si PSM systems at different institutions and mounting angle configurations, as reported in Table~\ref{tab:NRMSE}.

For tests involving sinusoidal motion, we generate the reference trajectory using Eq.~\ref{eq:sin_motion}.

\begin{equation}
    \begin{split}
        q^d_1 &= \alpha_{ori} \sin{(\omega_{gen} t)} \\
        q^d_2 &= \alpha_{ori} \sin{(\omega_{gen} t)} \\
        q^d_3 &= 0.12 + \alpha_{pos} \sin{(\omega_{gen} t)} \\
        q^d_4 &= q^d_5 = q^d_6 = q^d_7 = 0 \\
    \end{split}
    \label{eq:sin_motion}
\end{equation}
where $t$ denotes the runtime, $\alpha_{ori} = 20^\circ$ and $\alpha_{pos}=0.05\,\text{m}$ denote the amplitudes for the revolute and prismatic joints, respectively; and the frequency coefficient is set to $\omega_{gen} = 0.005\, rad/s$ (0.5 rad per time step at a 100\,Hz update rate).

\section{Results}

This section presents the results of experiments to assess the accuracy (Section \ref{sec:dyn-accuracy}) and computational cost (Section \ref{sec:runtime-perf}) of the predicted torque from the dynamics model, its performance for static gravity compensation (Section \ref{sec:grav-comp}) and dynamic trajectory tracking (Section \ref{sec:dynamic-tracking}), and to compare the two gravity compensation implementations (Section \ref{sec:gc-methods}).

The experimental results reported in Section~\ref{sec:grav-comp}, Section~\ref{sec:dynamic-tracking} and Section~\ref{sec:gc-methods} provide empirical validation of the proposed modeling, identification, and compensation framework under the evaluated configuration shown in Fig~\ref{fig:exp_setup}.


\subsection{Accuracy of Dynamics Model}
\label{sec:dyn-accuracy}

\begin{figure}[ht]
    \centering
    \includegraphics[width=0.95\linewidth]{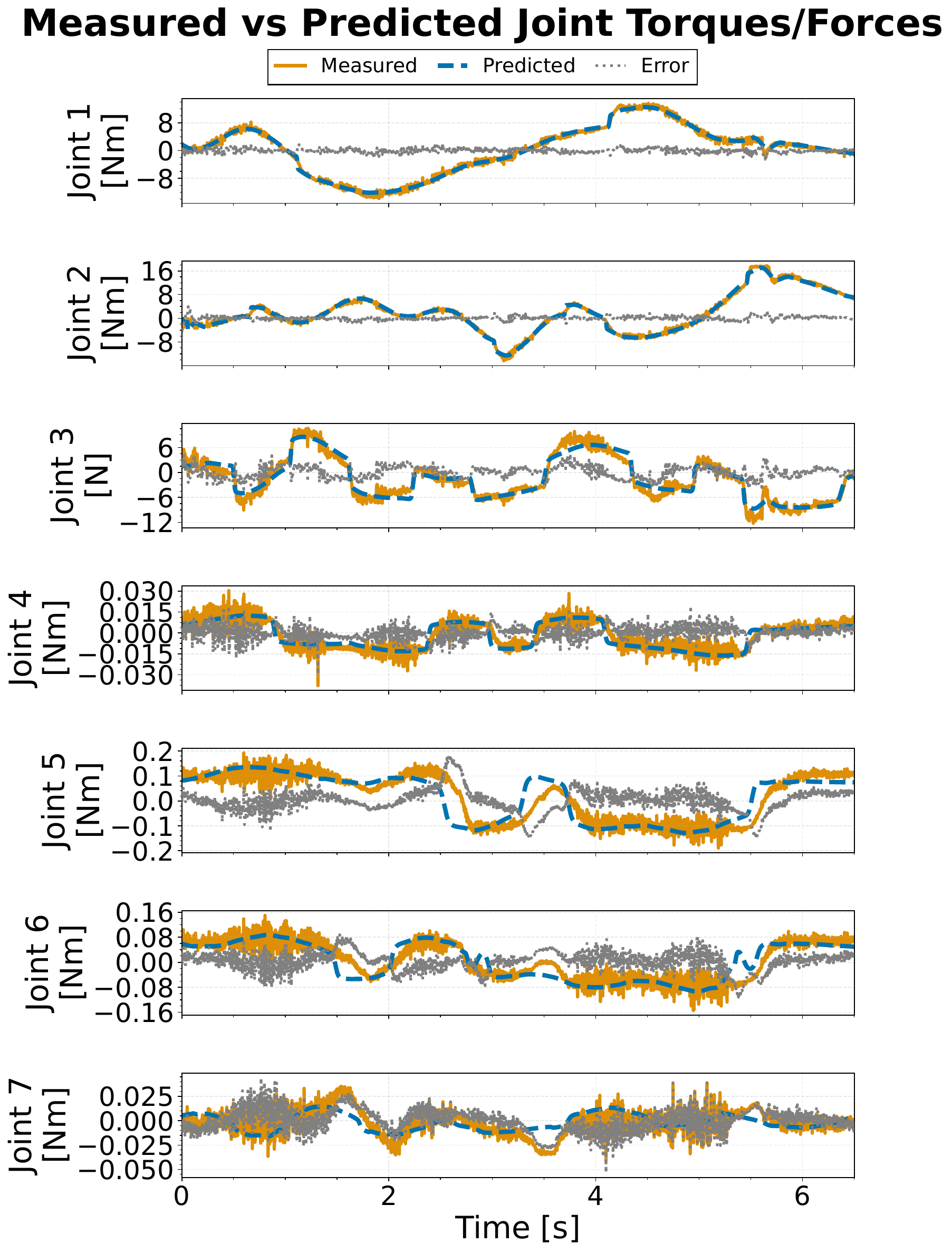}
    \caption{Measured and model-predicted joint torques/forces over a representative segment of an excitation trajectory.}
    \label{fig:sys_id_result}
\end{figure}

We evaluate the prediction accuracy of the identified dynamics model by comparing the model-predicted joint efforts $\bm{\hat{\tau}}$ with the corresponding measured joint efforts on independent excitation trajectories that were not used for parameter identification. Fig.~\ref{fig:sys_id_result} shows a representative segment of an excitation trajectory, where the predicted joint torques/forces closely follow the measured joint torques/forces for the primary load-bearing joints (joints 1-3). Quantitatively, we report the normalized root mean square error (NRMSE) computed over the full excitation trajectory, defined as:
\begin{equation}
    \begin{split}
        \text{NRMSE}(\%) = 100 \times \frac{\sqrt{\frac{1}{N}\sum_{k=1}^N{(\tau_k - \hat{\tau}_k)^2}}}{\sqrt{\frac{1}{N}\sum_{k=1}^N{\tau_k^2}}}
    \end{split}
    \label{eq:NRMSE}
\end{equation}
where $N$ denotes the number of points in the full excitation trajectory. To further assess the generalizability of the identification framework, we evaluate the parameter identification performance on multiple dVRK-Si PSM systems across different institutions and mounting-angle configurations. The resulting NRMSE values are summarized in Table~\ref{tab:NRMSE}.



\begin{table}[!b]
    \scriptsize
    \centering
    \caption{NRMSE Results for All Joints under 0$^\circ$ Horizontal Placement and 45$^\circ$ Downward-Tilted Surgical Scene Placement}
    \setlength{\tabcolsep}{4pt}
    \begin{tabular}{|r|rrrrrrr|}
        \hline
        Placement & J1 & J2 & J3 & J4 & J5 & J6 & J7 \\
        \hline
        0$^\circ$ Horizontal (JHU)
        & 7.43 & 10.57 & 21.91 & 36.28 & 45.29 & 45.49 & 87.56 \\
        \hline
        0$^\circ$ Horizontal (VU)
        & 10.72
        & 7.60
        & 14.55
        & 48.90
        & 31.99
        & 37.89
        & 87.04 \\
        \hline
        45$^\circ$ Downward (VU)
        & 9.45
        & 7.92
        & 20.51
        & 62.14
        & 40.01 
        & 64.49 
        & 68.47 \\
        \hline
    \end{tabular}
    \label{tab:NRMSE}
\end{table}

We can see that the first three joint torque/force predictions align closely with the measured values across the evaluated systems and mounting configurations. In contrast, the remaining joints exhibit higher NRMSE due to their substantially smaller torque magnitudes and high sensitivity to torque estimation noise, unmodeled transmission/friction effects, and larger velocity estimation errors due to lower encoder resolutions. The consistent prediction accuracy observed for all joints, especially for the primary load-bearing ones, across different systems and mounting configurations, suggests that the proposed identification framework has reasonable generalization capability for capturing the dominant dVRK-Si PSM dynamics. Nevertheless, the identified parameters and performance may still vary across hardware instances, instruments, and mounting conditions, motivating further large-scale validation in future work. Overall, these results indicate that the identified model captures the dominant PSM dynamics required for model-based compensation.

\subsection{Runtime Performance of Dynamics Model}
\label{sec:runtime-perf}

To evaluate the computational efficiency of the computed-torque term derived from the identified dynamic model, we measure the elapsed computation time for 10000 randomly generated joint configurations. The average computation time is 0.273\,ms with a standard deviation of 0.004\,ms on an AMD Ryzen\texttrademark\, 9 5950X CPU.

These results suggest that the inverse-dynamics evaluation using the identified model is computationally lightweight and can be performed within the time budget of low-level, real-time execution. In practice, the dominant bottleneck is the joint-state acquisition latency.

\subsection{Gravity Compensation Evaluation (Static)}
\label{sec:grav-comp}

To validate our gravity compensation results, we perform a drift test by first moving the PSM to a desired pose using its position controller (a PD controller) and then switching to the (open-loop) gravity-compensation-torque-based control model described in section~\ref{subsubsec: I-1} for 5 seconds. A 5\,s evaluation window was selected to provide sufficient time for gravity-induced drift to become observable while limiting the contribution of slower disturbances and maintaining a controlled open-loop test duration. The robot is considered to be drifting when moving more than 1 degree or 1 mm within the 5 seconds. We collect the kinematic data of the PSM and report the errors between the desired joint values and the measured joint values. We also determine the lower/upper bound torques (LB/UB $\tau$) which, due to static friction, are sufficient to hold the joint in position.  From Table~\ref{tab:psm_para}, we can see that gravity mainly influences the first three joints, and thus we only focus on those measurements and evaluations. We select five random poses for testing and the results are shown in Table \ref{tab:drift_test}.

\begin{table}[ht]
    \scriptsize
    \begin{center}
    \caption{Drift Test Results}
    \begin{tabular}{|c|@{}>{\hspace{2pt}}c<{\hspace{2pt}}@{}|r|r|r|r|r|r|}
        \hline
        \makecell{\textbf{Item}} & \makecell{\textbf{Pose} \\ \textbf{Idx}} & \makecell{\textbf{PD} \\ \textbf{Pos} \\ \textbf{Err}} & \makecell{\textbf{GC} \\ \textbf{Pos} \\ \textbf{Err}} & \makecell{\textbf{PD} \\ $\mathbf{\tau}$} & \makecell{\textbf{GC} \\ $\mathbf{\tau}$} & \makecell{\textbf{Non-} \\ \textbf{Drift} \\ \textbf{LB} \\ $\mathbf{\tau}$} & \makecell{\textbf{Non-} \\ \textbf{Drift} \\ \textbf{UB} \\ $\mathbf{\tau}$} \\
        \hline
        \multirow{5}{*}{\makecell{Joint1 \\ (deg \\or\\N$\cdot$m)}} & 1 & 0.14 & 0.04 & -1.42 & -1.07 & -0.11 & -1.76\\
        \cline{2-8}
        & 2 & 0.06 & 0.04 & 0.60 & 1.25 & 0.43 & 2.17\\
        \cline{2-8}
        & 3 & 0.85 & 0.16 & 8.87 & 7.67 & 6.06 & 9.75\\
        \cline{2-8}
        & 4 &0.50 & 0.11 & -5.25 & -6.39 & -4.66 & -7.99\\
        \cline{2-8}
        & 5 & 0.27 & 0.05 & -2.81 & -3.61 & -2.48 & -4.51\\
        \hline
        \multirow{5}{*}{\makecell{Joint2 \\ (deg \\or\\N$\cdot$m)}} & 1 & 0.50 & 0.13 & 5.29 & 4.32 & 2.98 & 5.75 \\
        \cline{2-8}
        & 2 & 0.45 & 0.12 & 4.67 & 5.73 & 3.84 & 7.34\\
        \cline{2-8}
        & 3 & 0.03 & 0.03 & 0.31 & 0.08 & -0.87 & 0.85\\
        \cline{2-8}
        & 4 & 0.03 & 0.02 & -0.32 & -1.44 & -0.18 & -2.64\\
        \cline{2-8}
        & 5 & 0.37 & 0.14 & -3.83 & -2.97 & -1.96 & -4.61\\
        \hline
        \multirow{5}{*}{\makecell{Joint3 \\ (mm \\or\\N)}} & 1 & 1.49 & 0.25 & -8.95 & -7.79 & -4.75 & -9.35\\
        \cline{2-8}
        & 2 & 0.36 & 0.13 & -2.19 & -4.41 & -2.07 & -5.96\\
        \cline{2-8}
        & 3 & 0.46 & 0.63 & -2.77 & 0.14 & -2.79 & 2.03\\
        \cline{2-8}
        & 4 & 1.05 & 0.35 & -6.31 & -4.56 & -1.83 & -6.67\\
        \cline{2-8}
        & 5 & 0.78 & 0.39 & 4.68 & 2.97 & 0.92 & 4.78\\
        \hline
        \hline
        \multicolumn{2}{|c|}{\textbf{Pose Idx}} & \textbf{x} & \textbf{y} & \textbf{z} & $\mathbf{r_x}$ & $\mathbf{r_{y}}$ & $\mathbf{r_{z}}$\\
        \hline
        \multicolumn{2}{|c|}{1} & 0.00 & 0 .00 & 113.50 & 180.00 & 0.00 & -90.00 \\
        \hline
        \multicolumn{2}{|c|}{2} & -123.93 & 101.14 & 108.87 & -126.83 & -20.18 & -65.27\\
        \hline
        \multicolumn{2}{|c|}{3} & -78.68 & -50.46 & -2.30 & -88.59 & -0.90 & -122.65\\
        \hline
        \multicolumn{2}{|c|}{4} & 50.10 & -45.76 & 48.66 & 129.09 & 22.45 & -64.83\\
        \hline
        \multicolumn{2}{|c|}{5} & 174.45 & 67.87 & -79.84 & 66.66 & 7.98 & -79.84\\
        \hline
    \end{tabular}
    \label{tab:drift_test}
    \end{center}
    \footnotesize{Note for the table: The units of the poses are mm and deg. The 6D pose coordinates are based on the frame $F_{RCM}$ shown in Fig.~\ref{fig:kinematics}. GC stands for gravity compensation.}
\end{table}

Results indicate that the gravity-compensation-torque-based control effectively maintains the robot's pose without noticeable drift. Furthermore, the gravity-compensation approach usually achieves a smaller steady-state error compared to the PD control. The joint torques generated by the gravity-compensation method for all three measured joints fall within their respective lower and upper bounds and, in most cases, are close to the midpoint of that range. For the third joint, some offset from the midpoint can be attributed to the significant friction in the prismatic joint and the cannula. Preliminary investigations also show that the gravity compensation can be generalized on different dVRK-Si PSMs with the same mounting-angle configuration.

\subsection{Dynamic Tracking Accuracy Evaluation}
\label{sec:dynamic-tracking}

\begin{figure}[!t]
    \centering
    \includegraphics[width=0.96\linewidth]{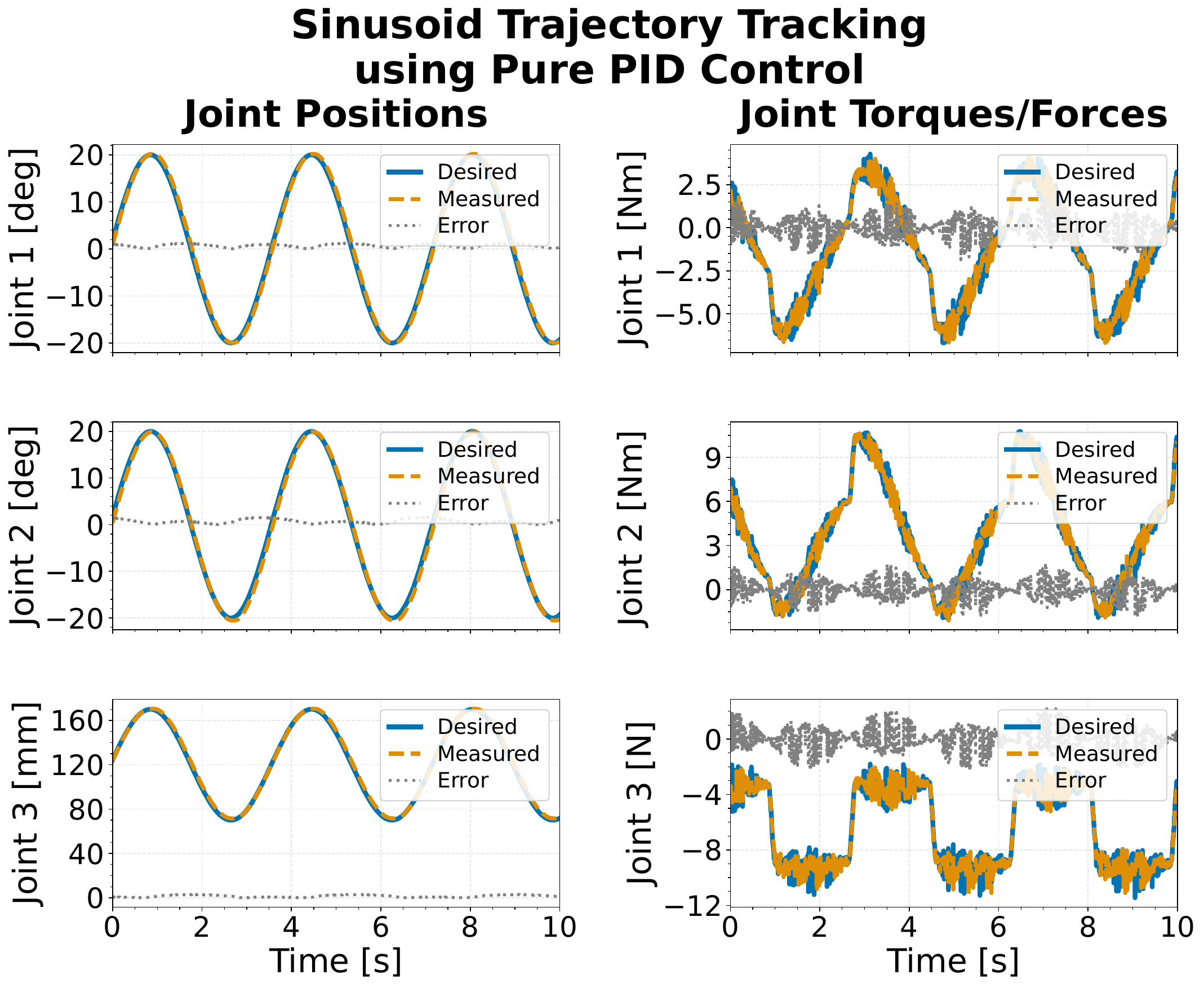}
    \includegraphics[width=0.96\linewidth]{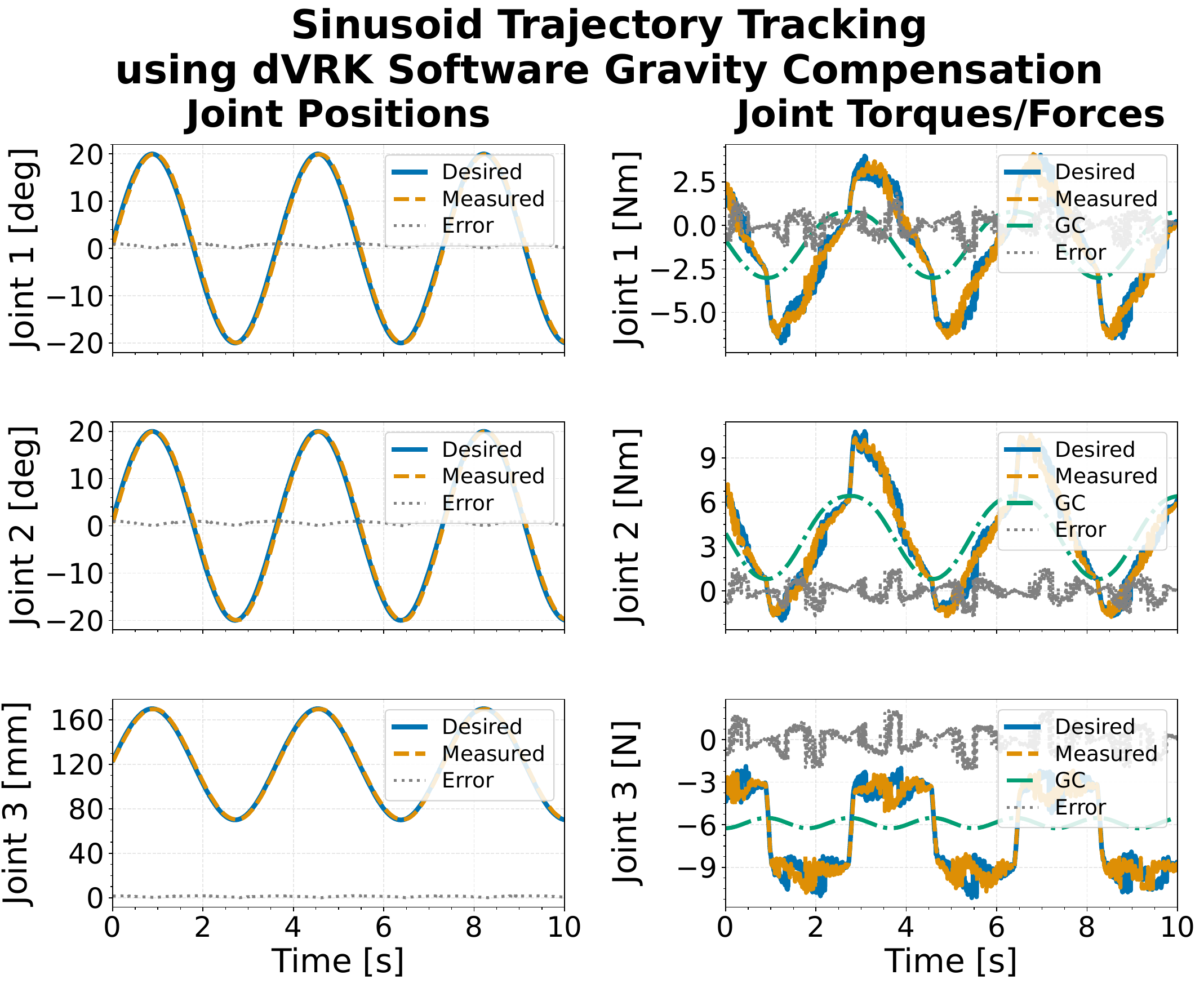}
    \includegraphics[width=0.96\linewidth]{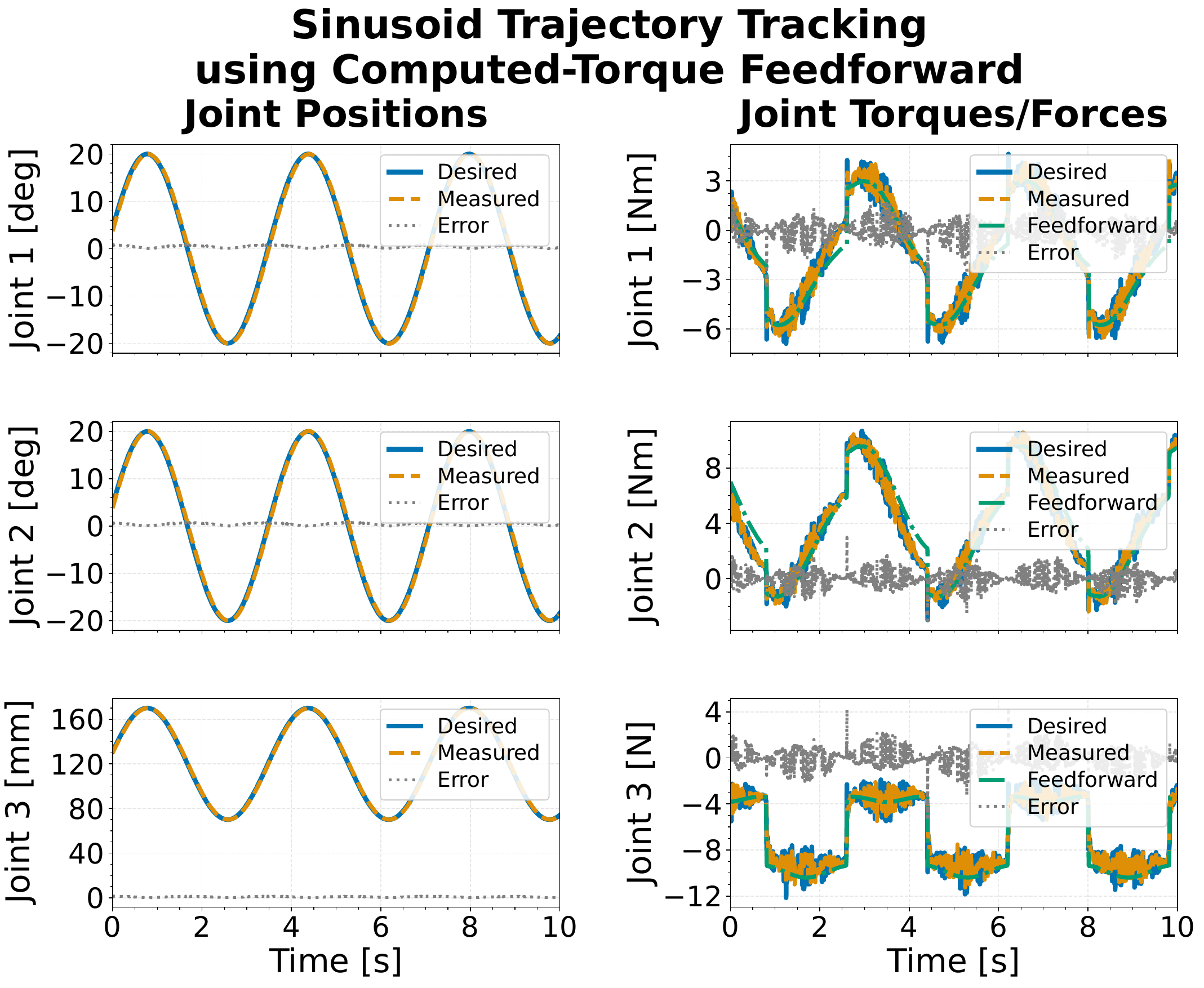}
    \caption{Sinusoidal trajectory tracking using PID controller only (top), PID controller with dVRK software gravity compensation (middle), and PID controller with computed torque feedforward (bottom). Left panels compare desired and measured joint positions and report the position tracking error $\bm{q}_{err} = \bm{q}_{des} - \bm{q}_{meas}$. Right panels show the corresponding torque/force, including the desired joint efforts, measured joint efforts, and the errors between the measured and desired joint efforts $\bm{\tau}_{err} = \bm{\tau}_{des} - \bm{\tau}_{meas}$ over a 10\,s trial.}
    \label{fig:compute_torque_result_all}
\end{figure}

We evaluate dynamic tracking performance by commanding the robot to track the sinusoidal reference trajectory in Eq. \eqref{eq:sin_motion} under three conditions: (1) PID control only, (2) PID control with dVRK software gravity compensation (Section \ref{sec:dvrk-gravity}), and (3) PID control with computed-torque feedforward (Section \ref{subsubsec: I-2}). The default dVRK-Si PID gains are used for all conditions. Results are shown in Fig.~\ref{fig:compute_torque_result_all} and tracking error statistics are summarized in Table~\ref{tab:track_err}.


\begin{table}[htbp]
    \footnotesize
    \begin{center}
    \caption{Tracking Error Statistics}
    \begin{tabular}{|c|@{}>{\hspace{2pt}}c<{\hspace{2pt}}@{}|c|c|c|c|}
        \hline
        \makecell{\textbf{Item}} & \makecell{\textbf{Joint} \\ \textbf{Idx}} & \makecell{\textbf{Mean}} & \makecell{\textbf{STD}} & \makecell{\textbf{RMSE}} & \makecell{\textbf{Max}} \\
        \hline
        \multicolumn{6}{|c|}{\textbf{Joint Position Errors} \quad (deg or mm)} \\
        \hline
        \multirow{3}{*}{\makecell{PID with \\  Torque \\ Feedforward}} & 1 & 0.46 & 0.24 & 0.52 & 0.94\\
        \cline{2-6}
        & 2 & 0.44 & 0.22 & 0.49 & 1.07 \\
        \cline{2-6}
        & 3 & 0.78 & 0.40 & 0.88 & 1.57 \\
        \hline
        \multirow{3}{*}{\makecell{PID with \\ dVRK Software \\ Gravity Compensation}} & 1 & 0.68 & 0.30 & 0.74 & 1.13 \\
        \cline{2-6}
        & 2 & 0.66 & 0.30 & 0.73 & 1.14 \\
        \cline{2-6}
        & 3 & 1.25 & 0.46 & 1.33 & 2.05 \\
        \hline
        \multirow{3}{*}{\makecell{PID only}} & 1 & 0.70 & 0.35 & 0.78 & 1.27 \\
        \cline{2-6}
        & 2 & 0.73 & 0.46 & 0.86 & 1.60\\
        \cline{2-6}
        & 3 & 1.35 & 1.02 & 1.70 & 3.11\\
        \hline
        \multicolumn{6}{|c|}{\textbf{Joint Torque Errors} \quad (Nm or N)} \\
        \hline
        \multirow{3}{*}{\makecell{PID with \\ Computed Torque \\ Feedforward}} & 1 & 0.48 & 0.38 & 0.62 & 3.57 \\
        \cline{2-6}
        & 2 & 0.48 & 0.37 & 0.61 & 3.05 \\
        \cline{2-6}
        & 3 & 0.70 & 0.53 & 0.88 & 5.18 \\
        \hline
        \multirow{3}{*}{\makecell{PID with \\ dVRK Software \\ Gravity Compensation}} & 1 & 0.49 & 0.38 & 0.62 & 2.06 \\
        \cline{2-6}
        & 2 & 0.49 & 0.38 & 0.61 & 2.10 \\
        \cline{2-6}
        & 3 & 0.71 & 0.53 & 0.88 & 2.78 \\
        \hline
        \multirow{3}{*}{\makecell{PID only}} & 1 & 0.49 & 0.38 & 0.62 & 2.93\\
        \cline{2-6}
        & 2 & 0.49 & 0.38 & 0.61 & 2.42\\
        \cline{2-6}
        & 3 & 0.70 & 0.52 & 0.87 & 4.05\\
        \hline
    \end{tabular}
    \label{tab:track_err}
    \end{center}
    \footnotesize{Note for the table: the joint position and torque errors are calculated by the difference between the desired (setpoint) and measured values acquired from the dVRK software, i.e. $\bm{q}_{err} = |\bm{q}_{des} - \bm{q}_{meas}|$ and $\bm{\tau}_{err} = |\bm{\tau}_{des} - \bm{\tau}_{meas}|$. \textbf{STD} and \textbf{RMSE} denote the standard deviation and root mean square error. Joints 1 and 2 are revolute joints and joint 3 is a prismatic joint.}
\end{table}

The results in Table~\ref{tab:track_err} show that the position tracking error is greatest in the PID only configuration, is somewhat reduced with gravity compensation (feedforward), and significantly reduced with the full computed-torque feedforward. 
Notably, this improvement is seen on all position tracking metrics (mean, STD, RMSE, and max), with the largest gains on the most gravity sensitive joint (the third joint).

\begin{figure}[ht]
    \centering
    \includegraphics[width=0.9\linewidth]{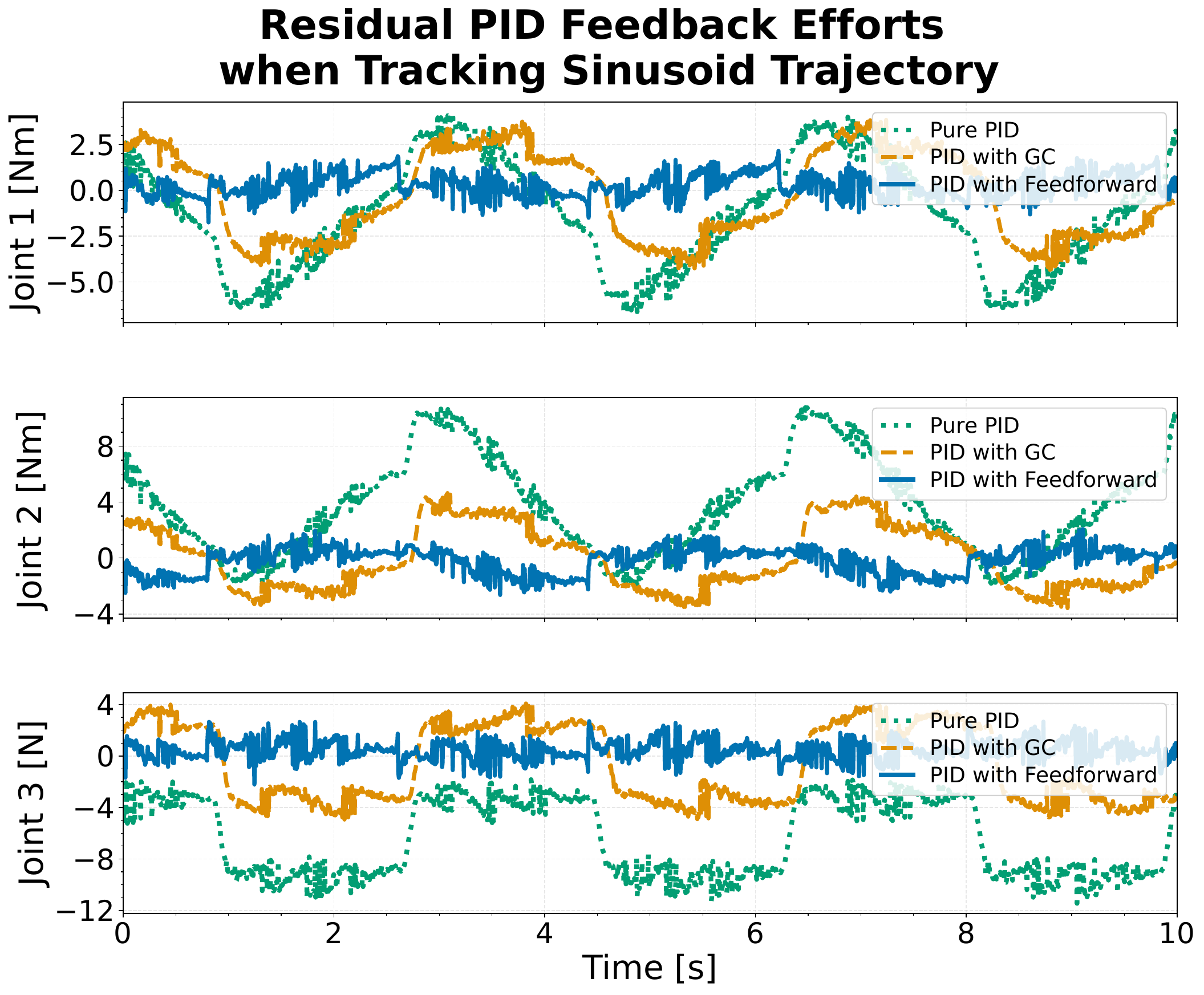}
    \caption{Residual PID feedback joint torque/force comparison; contributions due to gravity compensation and computed-torque feedforward are excluded.}
    \label{fig:pid_result}
\end{figure}

The torque tracking error is similar between the three conditions, as expected since the PID feedback controller compensates for the remaining torque required to attempt to follow the reference trajectory. This effect is illustrated in Fig.~\ref{fig:pid_result}, which shows the torque output of the PID controller. In the case of the PID only configuration, this is the total torque applied to the motor, but in the other two conditions it is the residual torque that is added to the gravity or computed-torque feedforward terms. As seen in Fig.~\ref{fig:pid_result}, the PID controller output with the computed-torque feedforward was much closer to zero than the other two configurations, indicating good accuracy of the dynamics model prediction. Interestingly, although most torque tracking metrics were similar, the maximum error was largest with the computed-torque feedforward. This may indicate that the computed-torque feedforward controller was more aggressive than the other two controllers, perhaps over-estimating the torque or exceeding actuator limitations.

\subsection{Comparison of Gravity Compensation Methods}
\label{sec:gc-methods}

\begin{figure}[ht]
    \centering
    \includegraphics[width=0.85\linewidth]{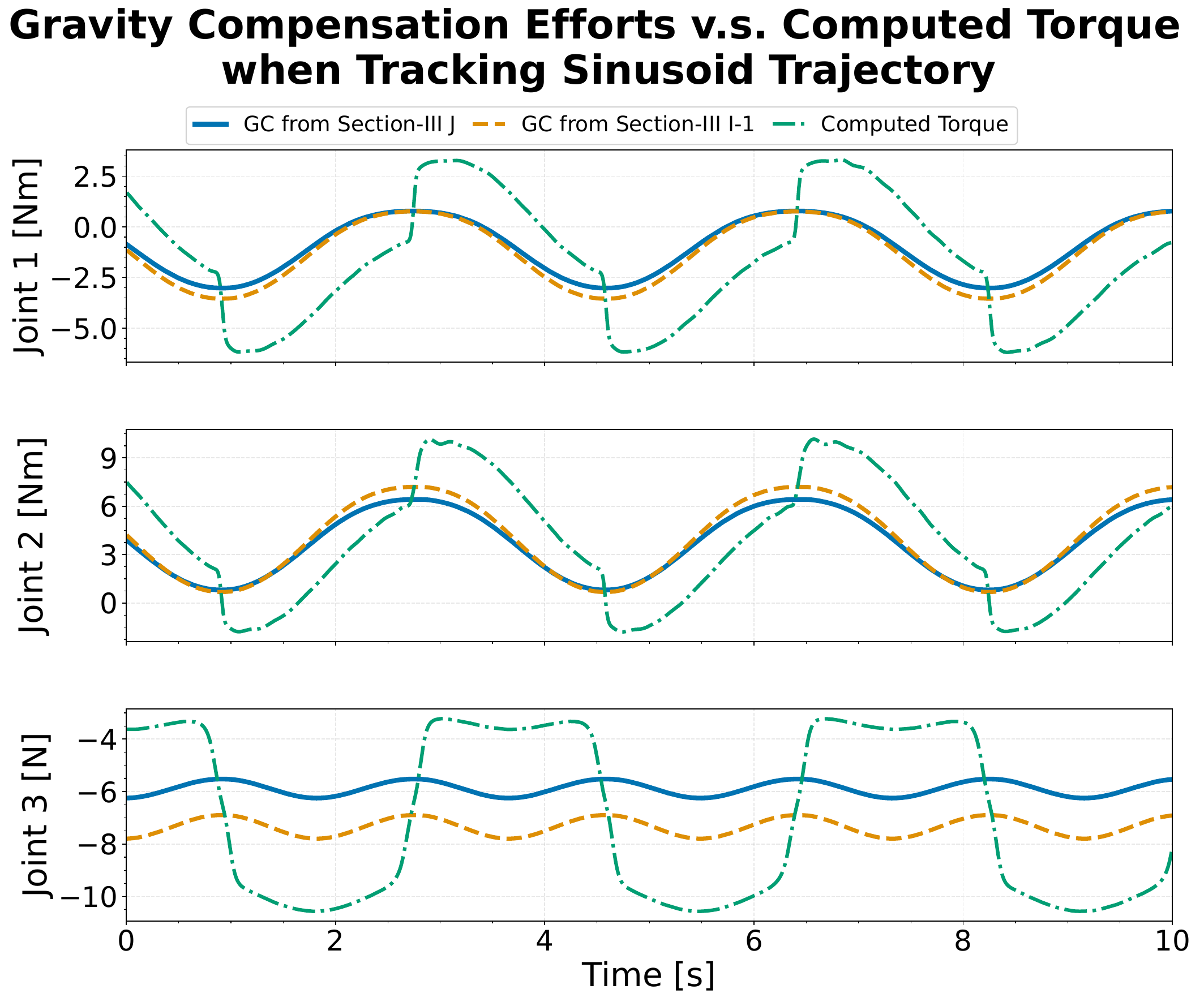}
    \caption{Comparison of two gravity-compensation implementations (from Section~\ref{subsubsec: I-1} and Section~\ref{sec:dvrk-gravity}) and the full computed-torque term over a sinusoidal trajectory.}
    \label{fig:dyn_track_result}
\end{figure}

Figure~\ref{fig:dyn_track_result} compares the two gravity compensation (GC) implementations while tracking the sinusoidal trajectory given by Eq.~\ref{eq:sin_motion}: (1) using the full dynamics model with velocity and acceleration set to zero (Section~\ref{subsubsec: I-1}), and (2) the simplified gravity compensation model currently included in the dVRK software (Section~\ref{sec:dvrk-gravity}).
In addition, the full computed-torque term is shown for reference. 
Both GC curves are smooth and periodic, which is consistent with a configuration-dependent gravity load, while the computed torque exhibits larger excursions and sharp transitions, reflecting additional motion-dependent dynamics (e.g., inertial/Coriolis terms) beyond gravity. The simplified gravity-compensation variant produces more conservative efforts because it ignores the friction and spring effects, reducing the risk of over-compensation.

\section{Discussion}

The experimental results validate that the identified dynamics model is sufficiently accurate for model-based compensation on the dVRK-Si PSM. In particular, gravity compensation based on the identified model substantially improves static holding performance by reducing steady-state joint errors and end-effector drift, and it also reduces the feedback nonlinearity on the baseline PID loop during motion. Furthermore, incorporating the full inverse-dynamics term as a feedforward component can improve trajectory tracking accuracy beyond gravity-only compensation, indicating that the model captures motion-dependent dynamics that are not addressed by gravity compensation alone.

A practical implication of these findings is that different levels of model complexity are appropriate for different integration points in the dVRK control stack. The full dynamic model enables computed-torque feedforward control and provides insight into torque limits and model prediction accuracy, while a simplified gravity-compensation variant can offer improved robot robustness for high-rate execution.

\subsection{Limitations}


All experiments in this manuscript are conducted on a single dVRK-Si PSM mounted at a fixed inclination (zero-degrees) relative to the table, as shown in Fig.~\ref{fig:exp_setup}. Although preliminary observations suggest that the behavior of the gravity compensation approach in section~\ref{subsubsec: I-1} generalizes across different PSMs with the same mounting-angle configuration, a broader evaluation across multiple mounting angles, instruments, and hardware instances is still needed to fully characterize generalization and repeatability.

The computed-torque feedforward implementation runs at the default dVRK-ROS software update rate (100\,Hz) and is currently limited by communication latency and the Python-based execution pipeline. These constraints can hinder low-level, real-time control at servo rates (e.g., full computed-torque control~\cite{siciliano2008springer}).

Furthermore, the current static and dynamic control evaluations provide descriptive empirical validation under a limited set of tested configurations, rather than statistically exhaustive evidence across the full workspace or diverse trajectory classes. Future work will include repeated trials, expanded static pose sets, multiple dynamic trajectories, and statistical analyses such as confidence intervals, effect sizes, and significance testing. These additional experiments will enable a more rigorous assessment of repeatability, robustness, and generalizability across different operating conditions.

\subsection{Future Work}

The current framework is implemented in Python with a ROS interface to the dVRK, but will be converted to a C++ implementation that is directly integrated with the dVRK software stack to enable lower-latency, higher-rate execution.
Some existing C++ toolkits, such as the Pinocchio package~\cite{carpentier2019pinocchio}, could be leveraged
for this effort.

Moreover, the PID gains used for sinusoidal trajectory tracking are kept identical across all control scenarios and are not re-tuned. As shown in Fig.~\ref{fig:dyn_track_result}, both gravity compensation and computed-torque feedforward reduce the residual PID feedback joint torques/forces. These results suggest that the PID gains should be fine-tuned for each scenario to achieve optimal tracking performance.

In addition, in this study, we only evaluate the effects of overall gravity compensation and computed-torque feedforward terms. Future work will further isolate the contributions of individual modeling and identification components, including physical feasibility constraints, optimized excitation trajectory design, smooth $\tanh$-based friction modeling, motor inertia, and spring terms, to provide a more detailed understanding of their respective effects on model accuracy and control performance.

Last but not least, the higher NRMSE prediction errors on low-torque joints could motivate enhanced modeling of friction, hysteresis, and transmission effects (e.g., joint-specific friction regimes, backlash, cable-driven transmission nonlinearities, and cannula/trocar interactions). Incorporating these effects (or identifying them separately) is expected to improve both inverse-dynamics prediction accuracy and the reliability of model-based compensation during slow motions and direction reversals.


\section{Conclusion}

This manuscript presented a complete modeling and identification framework tailored to the dVRK-Si PSM, including:
\begin{enumerate}
    \item a kinematic formulation that accounts for the closed-chain parallelogram mechanism;
    \item an Euler--Lagrange dynamics derivation expressed in linear-in-parameters form;
    \item a physically consistent identification pipeline driven by an optimized excitation trajectory.
\end{enumerate}

Using the identified parameters, we implemented real-time gravity compensation and computed-torque feedforward control using the dVRK software package. Experiments on a physical dVRK-Si PSM demonstrated that model-based gravity compensation substantially improves static performance by reducing steady-state errors and end-effector drift, while computed-torque feedforward control further improves trajectory tracking by compensating motion-dependent dynamics beyond gravity. Overall, these results establish a practical foundation for reliable model-based control, realistic simulation, and downstream autonomy research on the dVRK-Si platform.

\section*{Acknowledgement}
Thanks to Dale Bergman, Alessandro Gozzi and the Intuitive Foundation for all the hardware support of the dVRK-Si.

\bibliographystyle{IEEEtran}
\bibliography{references}

\end{document}